\documentclass[lettersize,journal]{IEEEtran}
\usepackage{amsmath,amsfonts}
\usepackage{algorithmic}
\usepackage{algorithm}
\usepackage{array}
\usepackage[caption=false,font=normalsize,labelfont=sf,textfont=sf]{subfig}
\usepackage{textcomp}
\usepackage{stfloats}
\usepackage{url}
\usepackage{verbatim}
\usepackage{graphicx}
\usepackage{cite}
\usepackage{lineno}
\usepackage{multirow}
\usepackage{placeins}
\usepackage{xcolor}

\begin{document}

\title{Deep Learning Methods for Adjusting Global MFD Speed Estimations to Local Link Configurations}


\author{Zhixiong Jin$^{1,2}$, Dimitrios Tsitsokas$^{2}$, Nikolas Geroliminis$^{2}$, and Ludovic Leclercq$^{1,\star}$
\thanks{$^{1}$Zhixiong Jin is with Univ. Gustave Eiffel, ENTPE, LICIT-ECO7, Lyon, France, and also with École Polytechnique Fédérale de Lausanne (EPFL), Urban Transport Systems Laboratory (LUTS), Lausanne, Switzerland  
{\tt\small zhixiong.jin@univ-eiffel.fr}}%
\thanks{$^{2}$Dimitrios Tsitsokas with École Polytechnique Fédérale de Lausanne (EPFL), Urban Transport Systems Laboratory (LUTS), Switzerland  
{\tt\small dimitrios.tsitsokas@epfl.ch}}%
\thanks{$^{2}$Nikolas Geroliminis with École Polytechnique Fédérale de Lausanne (EPFL), Urban Transport Systems Laboratory (LUTS), Switzerland  
{\tt\small nikolas.geroliminis@epfl.ch}}%
\thanks{$^{1}$Ludovic Leclercq with Univ. Gustave Eiffel, ENTPE, LICIT-ECO7, Lyon, France
{\tt\small ludovic.leclercq@univ-eiffel.fr}}%
\thanks{$^{\star}$Corresponding author.}
\thanks{This project has received funding from the European Union's Horizon 2020 research and innovation program under Grant Agreement no. 953783 (DIT4TraM)}
}



\maketitle

\begin{abstract}
In large-scale traffic optimization, models based on Macroscopic Fundamental Diagram (MFD) are recognized for their efficiency in broad network analyses. However, they fail to reflect variations in the individual traffic status of each road link, leading to a gap in detailed traffic optimization and analysis. To address the limitation, this study introduces a Local Correction Factor (LCF) that represents local speed deviations between the actual link speed and the MFD average speed based on the link configuration. The LCF is calculated using a deep learning function that takes as inputs the average speed from the MFD and the road network configuration. Our framework integrates Graph Attention Networks (GATs) with Gated Recurrent Units (GRUs) to capture both the spatial configurations and temporal correlations within the network. Coupled with a strategic network partitioning method, our model enhances the precision of link-level traffic speed estimations while preserving the computational advantages of aggregate models. In our experiments, we evaluate the proposed LCF across various urban traffic scenarios, including different levels of origin-destination trip demand and distribution, as well as diverse road configurations. The results demonstrate the robust adaptability and effectiveness of the proposed model. Furthermore, we validate the practicality of our model by calculating the travel time of each randomly generated path, achieving an average error reduction of approximately 84\% relative to MFD-based results.

\end{abstract}


\section{Introduction}\IEEEPARstart{T}{he} effective management of large-scale urban traffic systems is critical to mitigate congestion, enhance mobility of road users, and ensure the sustainability of the urban environment. Within this scope, aggregated traffic models based on the Macroscopic Fundamental Diagram (MFD) have emerged as powerful tools in large-scale congestion management strategies. The MFD reveals the relationship between network efficiency (such as network-mean speed and travel production) and network accumulation under homogeneous traffic conditions \cite{geroliminis2008existence}. This relationship simplifies the analysis of traffic networks and enhances computational efficiency in solving traffic management and control optimization problems. Over the past decades, the MFD-based traffic models have informed various large-scale traffic management strategies, such as pricing schemes \cite{zheng2016time,yang2019heterogeneity,zheng2020area}, perimeter control \cite{fu2021perimeter,aalipour2018analytical,batista2021role,tsitsokas2023two,jiang2023hybrid,johari2021macroscopic,haddad2021traffic,chen2022data}, and multi-modal transportation network optimization \cite{wei2020modeling,loder2022optimal,beojone2021inefficiency,balzer2023dynamic}.

One of the significant challenges in applying MFD-based traffic models is their limited ability to reflect the variations of individual traffic states of each road link. These models typically assume uniform speed across all road sections within a region. This is particularly true for trip-based MFD approaches \cite{mariotte2017macroscopic,lamotte2018morning,balzer2022modal}, which on the one hand provide an individual and dynamic trip description over the travel paths, on the other hand, stick to average regional speed, making all links crossing similar. While this assumption benefits broad-scale analysis and optimization, it overlooks the unique spatial characteristics of individual road links, such as lane counts, presence of dedicated bus lanes, and topology. 
 As a result, these approaches are less effective for optimizing tasks that require detailed network configuration. For example, realistic route choice relies on accurate speed differentiation for each link, while dedicated bus-lane allocations—which reduce general vehicle-lane capacity and alter link speeds—require precise link-level speed estimations. The assumption of uniform speed in the MFD-based models fundamentally limits their applicability in tasks aimed at enhancing traffic efficiency through detailed, configuration-specific optimization strategies.

To address the limitations that might be important for some specific applications, our research introduces a novel approach: Local Correction Factor (LCF). The proposed concept precisely estimates the speed of individual links by integrating the network mean speed, which can be derived from the speed MFD, along with comprehensive network configuration elements, including the spatial attributes of each link and the network topology. By introducing the LCF, our approach preserves the computational efficiency of MFD‐based traffic models while accounting for configuration‐specific effects (e.g., allocation of dedicated bus lanes) that directly affect link performance. Even though fully detailed traffic simulations can model these effects accurately, they incur substantial computational expense when evaluating numerous layout scenarios. In contrast, the proposed approach achieves comparable accuracy at a small fraction of the cost, making it particularly well suited for strategic optimization tasks that depend on detailed network configurations.

In recent years, deep learning methods have emerged as powerful tools to solve complex problems by extracting information across various domains, including computer vision and natural language processing \cite{achiam2023gpt,szegedy2017inception}. These methods have also shown remarkable success in transportation engineering, such as trajectory analysis \cite{choi2018network,choi2019attention}, traffic prediction \cite{zheng2020gman,hwang2022asymmetric,zhang2019spatial}, and traffic safety studies \cite{jin2022deep,jin2023prediction}. Specifically, deep learning has demonstrated efficiency in traffic speed estimation and prediction, with numerous research validating its performance. For example, \cite{lv2014traffic} employs a Stacked Autoencoder (SAE) to learn low-dimensional representations of spatial–temporal traffic patterns for improved flow prediction. Similarly, \cite{zhang2019trafficgan} proposes TrafficGAN, a Generative Adversarial Network (GAN)-based model that generates realistic future traffic scenarios conditioned on historical data.

The challenge shifts towards strategically applying deep learning to leverage network configuration and its average speed data for LCF development. This requires a sophisticated understanding of both spatial and temporal dimensions of traffic data. Spatial embedding techniques capture the unique characteristics and interconnections of the road network, while temporal embedding accounts for the dynamic changes of the road network over time. These embeddings provide a comprehensive framework that enhances our model's ability to accurately estimate individual link speed. A review of the literature reveals an increasing interest in employing spatial and temporal embeddings to improve traffic estimation or prediction models. \cite{yu2017spatiotemporal} proposes Spatiotemporal Recurrent Convolutional Networks (SRCs) for speed prediction in urban areas. The model combines Deep Convolutional Neural Networks (DCNN) and Long Short-term Memory (LSTM) networks to capture the spatial dependencies of the network and the temporal dynamics of historical data. Similarly, \cite{cao2020cnn} develops a model based on CNN and LSTM models to extract the features of the daily and weekly periodicity of traffic speed. \cite{ma2022novel} uses CNNs and Gated Recurrent Units (GRUs) to extract deep features from the reconstructed spatial-temporal matrix. Even though the mentioned methods have achieved good results by considering spatial-temporal features of traffic data, these models often fail to fully incorporate the network topology information due to the restrictions of CNN models. In other words, the grid-structured input data limits the performance of the models. To address this issue, Graph Neural Networks (GNNs), which are designed to handle graph-structured data, have emerged as a promising alternative. GNNs effectively capture complex relationships and dependencies within the network topology, offering a more nuanced understanding of traffic dynamics. For example, \cite{zhao2019t} introduces a Traffic Graph Convolutional Network (T-GCN) model that integrates GCNs with GRUs to simultaneously process spatial relationships and temporal dependencies within traffic networks. This hybrid approach allows for a more accurate representation of traffic flow patterns, leveraging the graph structure of road networks to enhance prediction accuracy. \cite{li2017diffusion} proposes a Diffusion Convolutional Recurrent Neural Network (DCRNN), which comprises hybrid GCN models that capture the spatial dependency with the random walk and temporal dependency with GRUs. Similarly, \cite{zhang2019spatial} develops a Spatial-Temporal Graph Attention Network (ST-GAT) where GAT is used to extract the spatial dependencies of road networks, and then LSTM is used to extract the temporal features of speed data. \cite{zheng2020gman} builds the Graph Multi-Attention Network (GMAN) for long-term traffic prediction. GMAN employs an encoder-decoder architecture: the encoder compresses historical traffic data into a low-dimensional vector that preserves the key spatio-temporal patterns, and the decoder uses this vector to predict future speeds. Both modules incorporate stacked spatio-temporal attention blocks, where spatial attention dynamically weights the influence of different sensors on the road network and temporal attention captures non-linear dependencies across time. A key element is the transform attention layer, which uses scaled dot-product attention to directly map encoded historical features to future time steps, thereby reducing error propagation.
To further improve the performance of the model, in \cite{hwang2022asymmetric}, asymmetric characteristics of the forward and backward waves are considered in spatial-temporal embedding in GMAN. This exploration into spatial-temporal embeddings not only highlights their crucial role in traffic prediction models but also sets the stage for our proposed methodology, designed to advance speed estimation tasks for urban road networks.

This paper introduces a novel approach that extends beyond the current state-of-the-art by integrating the network mean speed with the spatial configuration of the urban road network to estimate the speed of individual links. The proposed methodology, which we name LCF, bridges the gap in accurately representing urban traffic speed heterogeneity at a granular level, maintaining the computational efficiency of aggregated traffic models. Our methodology leverages GATs to effectively capture the complex spatial dependencies within the urban network. GATs are selected for their ability to adaptively assign weights to neighboring nodes through attention mechanisms, enabling the model to focus on the most relevant spatial relationships. In the model, the attention mechanism calculates a score for each neighbor based on its feature similarity to the target node. These scores serve as weights, with more similar neighbors receiving higher values and exerting a greater influence on the target node's update. This capability is essential in urban road networks, where the significance of connections between nodes can vary substantially. Additionally, we employ GRUs to capture temporal correlations associated with congestion dynamics.GRUs are a streamlined variant of RNNs that employ two gating mechanisms—an update gate to regulate how much past information to retain, and a reset gate to control the incorporation of new inputs into the stored state. This architecture captures temporal dependencies with significantly fewer parameters than LSTM networks, yielding comparable predictive performance while reducing training time.

By integrating GATs and GRUs, our model effectively captures both the spatial and temporal features of traffic data, thereby enhancing link-specific speed estimations. Furthermore, we resort to network partitioning to enhance model performance.

Our contributions are summarized as follows: 

\begin{itemize}
    \item We propose the for accurately estimating the speed of individual links in urban road networks, by integrating network configuration information with the network mean speed provided by the MFD. This approach overcomes the limitations of traditional MFD-based models by offering a granular analysis that considers the spatial characteristics of individual road links.
    
    \item We introduce a novel deep learning framework that combines GATs with GRUs. This framework significantly improves the accuracy of the speed estimation by exploring the complex configuration of urban road networks and capturing the dynamics of traffic flow.
    
    \item We use a network partitioning method to divide the urban road network into more homogeneous regions to further enhance estimation accuracy. Comprehensive evaluations under varied traffic conditions—including different demand levels, origin-destination distributions, and road configurations—demonstrate the model's adaptability and effectiveness in diverse urban scenarios. This robust testing underscores its practicality for traffic optimization and management applications.
    
    
\end{itemize}


\section{Methodology} 
\subsection{Problem Definition}\label{subsec:problem definition}
\subsubsection{{\color{red}Road‑Network Representation}}
\begin{figure}[t]
    \centering
  \includegraphics[width=0.4\textwidth]{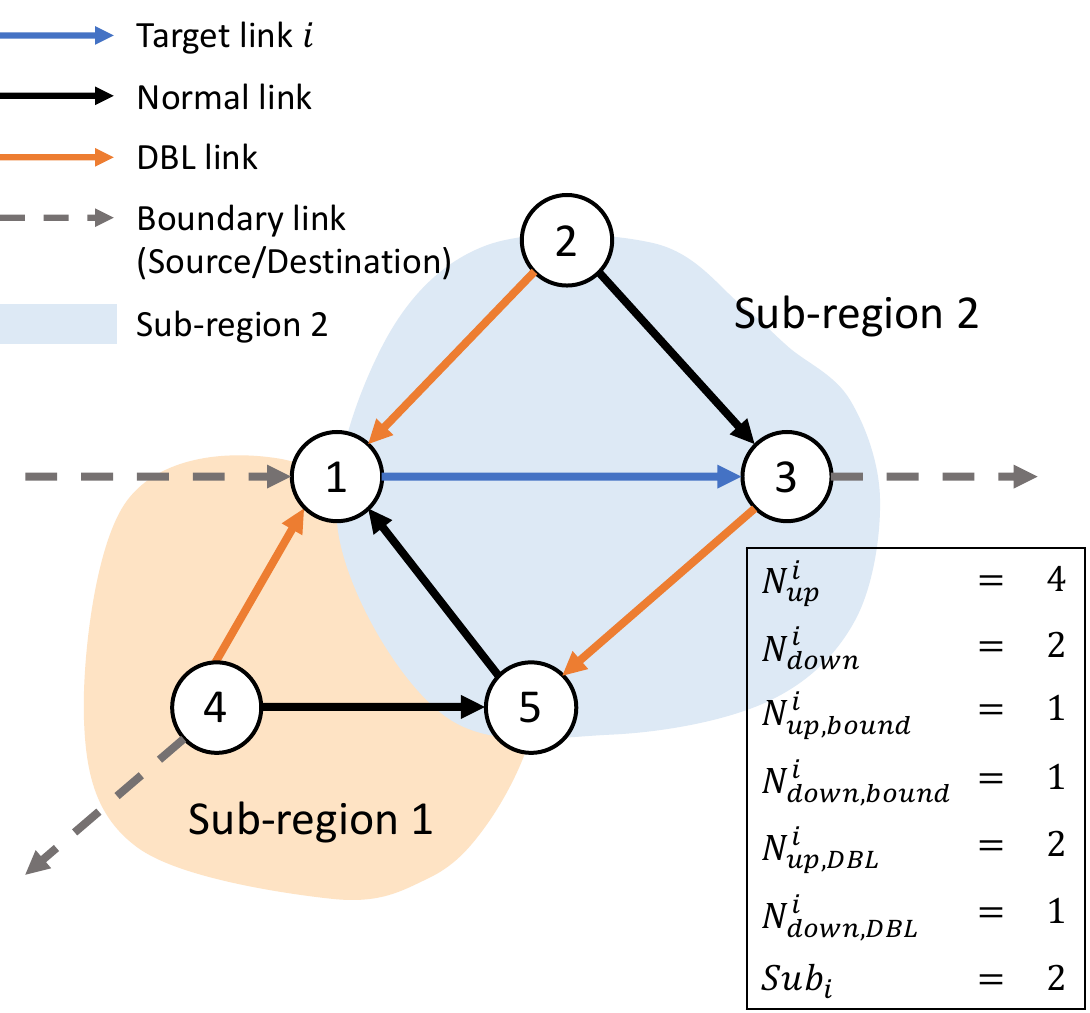}
  \caption{{\color{red}Illustration of additional road link attributes}}\label{fig 2:variable illustration}
\end{figure}

The road network is represented by a graph \(G = (\mathcal{V},E)\), where each physical link is modeled as a node to suit GNN feature processing. In the graph $G$, the vertex set $\mathcal{V}$ encapsulates spatial attributes of the network containing $N$ nodes, expressed as $\mathcal{V} = \{v_1, v_2, \ldots, v_N\}$. Each element $v_i$ contains essential node attributes, including the length of the road section $L_i$, the total number of lanes $N_{\text{total}}^i$, and the presence of dedicated bus lane $N_{\text{DBL}}^i$, with $v_i = (L_i, N_{\text{total}}^i, N_{\text{DBL}}^i) \in {\color{red}\mathcal{V}}$.  It is important to note that \( N_{\text{DBL}}^i \) serves as a dynamic attribute in our model. Assigning \( N_{\text{DBL}}^i \) a value of 0 or 1 for each node allows for the addition or removal of dedicated bus lanes on specific links (by conversion of one of the general purpose lanes of the link to dedicated bus lane), which enables the simulation of diverse road network scenarios without altering the network's fundamental topology. To further refine our model's spatial analysis capabilities, we integrate additional attributes for each node: the number of upstream (\(N_{\text{up}}^i\)) and downstream (\(N_{\text{down}}^i\)) links, the number of upstream (\(N_{\text{up,bound}}^i\)) and downstream (\(N_{\text{down,bound}}^i\)) links connected to network boundary, the presence of dedicated bus lanes in the upstream (\(N_{\text{up,DBL}}^i\)) and downstream (\(N_{\text{down,DBL}}^i\)) directions, and the label of sub-regions (\(Sub_i\)) (which will be discussed in detail later).  
{\color{red}Figure~\ref{fig 2:variable illustration} illustrates the details of these spatial attributes.}
{\color{red}The set \(E\) consists of edges \((v_i, v_j)\) representing connectivity between node pairs whose corresponding physical links share a junction, thereby illustrating the network’s physical linkages.}

\subsubsection{{\color{red}Speed Estimation}}
The objective of this study is to estimate the speed of individual road links at time \( t \) using the road network graph \( G \) and the network mean speed \( V_{\text{mean}}^{t} \) derived from speed MFD. This estimation is performed by calculating the LCF through {\color{red}deep learning models} \(f_\text{LCF}\), {\color{red} expressed as following equation},
\begin{linenomath}
\begin{equation}\label{eq:1}
\text{LCF}^{t} = f_{\text{LCF}}(G; V_{\text{mean}}^{t})
\end{equation}
\end{linenomath}
The estimated speeds of the individual road links are then obtained by adding the LCF to the network mean speed, as follows,
\begin{align}
V_{i}^t &=  V_{\text{mean}}^{t} + \text{LCF}_{i}^t
\end{align}
\noindent
where \( \text{LCF}_{i}^t \) denotes the correction factor for the \( i^{\text{th}} \) road link at time \( t \); \( V_{i}^t \) represents the estimated speed of the \( i^{\text{th}} \) road link. For practical reasons, we set up the model to directly estimate the link speeds as outputs.

\begin{figure*}[t]
    \centering
  \includegraphics[width=1\textwidth]{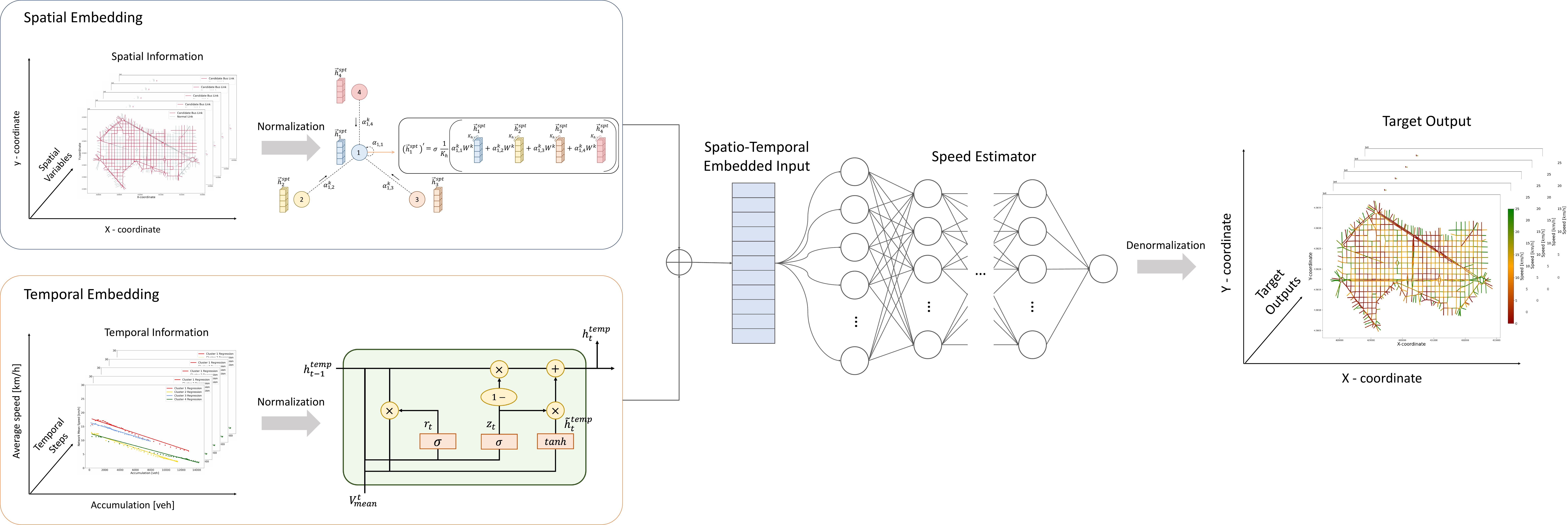}
  \caption{The model architecture of proposed LCF}\label{fig 1: model architecture}
\end{figure*}

\subsection{Overview}
This section outlines the development of the LCF, calculated using deep learning models to estimate the speed of each link in the road network. Our proposed model comprises three primary components: Spatial Embedding, Temporal Embedding, and Speed Estimation, as illustrated in Fig.~\ref{fig 1: model architecture}. Specifically, we employ {\color{red}GATs} to capture the network configuration and extract spatial features. Simultaneously, {\color{red}GRUs} are applied to analyze the network mean speed over time. To further improve the model's performance, we incorporate a network partitioning method. The final estimated speed results are obtained through fully connected layers by combining the two types of extracted inputs (Spatio-Temporal Embedded Input). The {\color{red} following subsection} will provide a detailed explanation of the data preprocessing, along with how we handle spatial and temporal data, and how the selected traffic simulator provides the required traffic data.

\subsection{{\color{red}Model Framework}}\label{subsec:method}

\subsubsection{Data Preprocessing and Network Partitioning}

The effectiveness of our proposed model depends on the meticulous preprocessing of the input data and the network partitioning procedures. These initial steps are crucial to ensure data consistency, improve model training efficiency, and address heterogeneity problems in urban road networks.

\paragraph{Data Preprocessing}

To build the LCF, we normalize the input variables to facilitate consistent and efficient model training. This involves using the min-max normalization technique to scale each component of the input variables within a uniform range (0 to 1). The process ensures that all features are equally represented, helping to prevent biases in the model.


\begin{linenomath}
\begin{equation}\label{eq:2}
x_{\text{norm}} = \frac{{x - \text{min}(x)}}{\text{max}(x) - \text{min}(x)}
\end{equation}
\end{linenomath}

\paragraph{Network Partitioning}

The urban road network exhibits significant heterogeneity, characterized by diverse traffic patterns and spatial features, which may reduce the accuracy of the MFD simulations \cite{ramezani2015dynamics}. Therefore, this study integrates a network partitioning technique aimed at identifying more homogeneous sub-regions, which can facilitate accurate speed estimation. 

Leveraging the requirement for network partitioning, we employ clustering methods to effectively segment the road network. This approach has been supported by extensive literature that highlights the efficiency of its usage in network partitioning \cite{lin2020road,chen2022urban,saeedmanesh2016clustering,lopez2017revealing,saeedmanesh2017dynamic}. While many advanced algorithms focus on optimizing network partitions based on compactness, connectivity, and traffic homogeneity, crucial for traffic management and control applications, our primary goal is to investigate whether network partitioning can improve speed estimation, rather than to identify the optimal partitioning algorithm. Therefore, we implemented the classical K-Means clustering algorithm by considering both the distance between links and the speed difference in the similarity metric. {\color{red} We considered two primary inputs—the geographical positions of road links and the mean speeds of each link during the period of peak network production—to jointly capture spatial layout and operational efficiency.} The geographical position of each link, denoted as \( \text{Loc}_i = (x_i, y_i) \), is determined by its midpoint coordinates. The mean speed of each link, \( V_i^{\text{mean}} \), is calculated over a defined time window \( T_w \) centered around the time of maximum production \( T_{\text{max}} \), specifically within the interval \( [T_{\text{max}} - T_w, T_{\text{max}} + T_w] \). To integrate these spatial and temporal data points for clustering, we construct a feature vector \( \mathbf{S}_i \) for each road link \( i \) by combining the geographical coordinates and the average speed through a weighted sum, as shown in {\color{red} following equation},

\begin{equation}\label{eq:3}
{\mathbf{S}_i} {=} 
{\begin{bmatrix}
\alpha \cdot x_i \\
\alpha \cdot y_i \\
\beta \cdot V_i^{\text{mean}}
\end{bmatrix},} {\quad  i = 1, 2, \dots, N}
\end{equation}
\noindent
here, \( \alpha \) and \( \beta \) are respective weights assigned to the spatial and temporal components to achieve a balanced representation of both factors in the clustering process. {\color{red}In this research, we choose $\alpha$ = 1, and $\beta$ = $\sqrt{2}$ to balance the contributions of the two spatial dimensions against the temporal feature in the K‑Means clustering process}. Furthermore, we define the time window, $T_{w}$, as 2 time steps, and the maximum production time, $T_{max}$, occurs at time step 40 based on the simulation settings. The set of all feature vectors is denoted as \( \mathbf{S} = \{ \mathbf{S}_1, \mathbf{S}_2, \dots, \mathbf{S}_N \} \).

Subsequently, we apply the K-Means algorithm to the feature set \( \mathbf{S} \) to partition the road links into \( K_c \) distinct clusters, where \( K_c \) represents the predetermined number of clusters (sub-regions). The clustering process is mathematically expressed as,

\begin{equation}
\{ r_k \}_{k=1}^{K_c} = \text{KMeans}(\mathbf{S}, K_c)
\end{equation}
\noindent
where \( r_k \) denotes the subset of road links assigned to cluster \( k \). To determine an appropriate number of clusters, we calculate the Within-Cluster Sum of Squares (WCSS) for various values of \( k \) and identify the “elbow” point—where the rate of decrease in WCSS sharply changes—indicating a meaningful partition of the data based on geographical positions and average speeds. From this analysis, we determined \( K_c = 4 \), which also aligns with the relatively symmetric, grid-like structure of the Barcelona road network. In this study, network partitioning was performed only once, based on a randomly selected scenario, and was consistently applied across all simulations. This consistent partitioning allows us to evaluate the impact of partitioning on speed estimation without introducing variability from changing cluster configurations.



This data normalization and network partitioning are crucial not only for preparing the input data for model training but also for addressing the inherent heterogeneity problem of urban road networks. By implementing these steps,  we establish a foundation for the development of accurate speed estimation models.

\subsubsection{Spatial Embedding} 
The spatial embedding plays a crucial role in capturing the spatial characteristics of the road network that significantly affect speed estimation. To deal with the complex spatial patterns of urban road networks, GNNs are ideal due to their ability to handle non-Euclidean data. {\color{red} Euclidean data consist of elements arranged on a regular grid in d-dimensional real space \((\mathbb{R}^d)\) (e.g., images on a 2-D pixel grid), where distances are measured with the standard Euclidean metric. In contrast, non-Euclidean data, like road networks, molecular structures, or social networks, have irregular connections between nodes.} This aligns perfectly with the non-Euclidean layout of road networks, facilitating precise modeling of intricate interactions and dependencies among the various links and nodes in the network.

\begin{figure}[t]
\centering
  \includegraphics[width=0.48\textwidth]{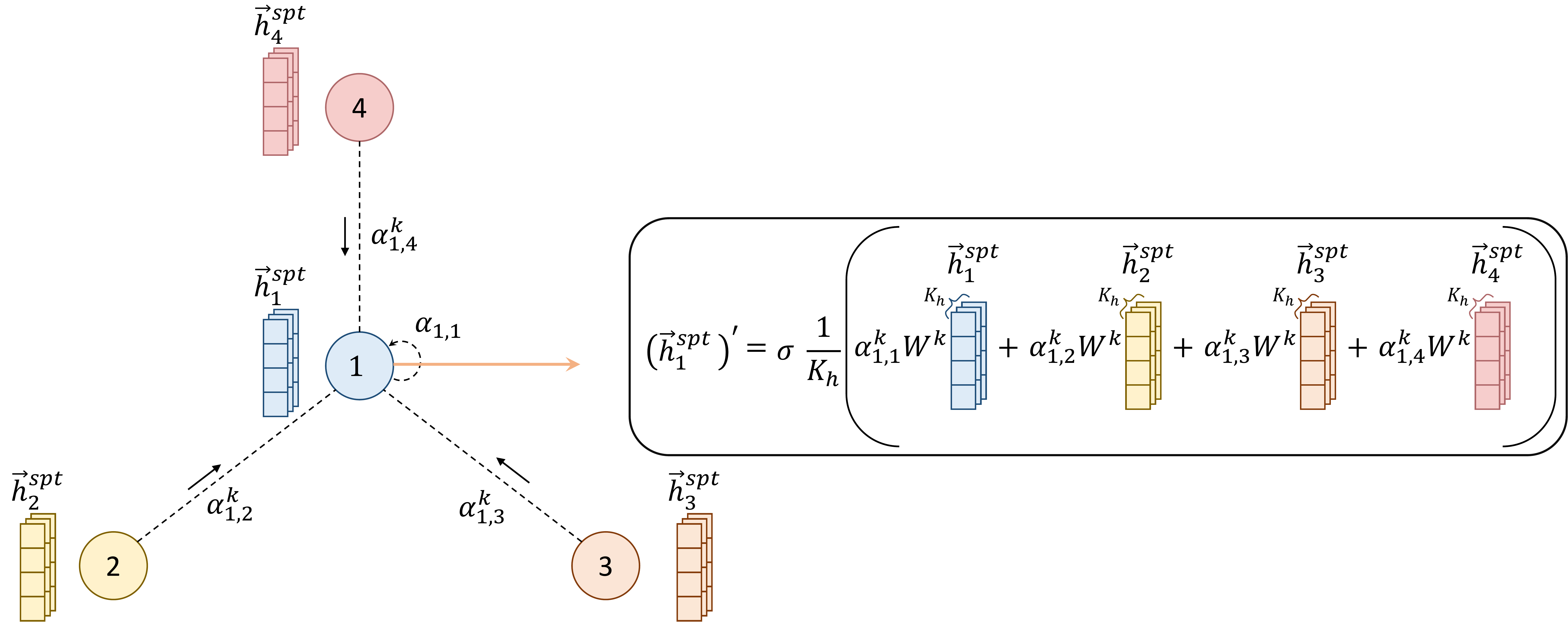}
  \caption{The architecture of the GATs}\label{fig:GAT}
\end{figure}

Within the wide range of GNN architectures, we use the GAT to further refine our spatial analysis. The selection of the GAT is motivated by its unique attention mechanism, which dynamically assigns weights to the connections between nodes in the network \cite{velivckovic2017graph}. This functionality proves especially beneficial in urban road networks, where the importance of connections between nodes can significantly vary due to diverse factors such as connectivity, distance, and prevailing traffic conditions. 
In this study, the process of GATs to extract spatial features can be summarized into the following three steps: Initialization, Attention coefficients calculation, and Multi-head attention calculation. The choice of activation functions is based on the work from \cite{velivckovic2017graph} {\color{red}and the related process is depicted in Fig.~\ref{fig:GAT}.}

\noindent

\paragraph{Initialization} Firstly, we concatenate multiple features of nodes and represent the inputs as \( \vec{h}_1^{spt}, \vec{h}_2^{spt}, \ldots, \vec{h}_N^{spt} \), $\vec{h}_i^{spt} \in \mathbb{R}^F$, where $N$ is the number of nodes (or number of the links in the road network), and $F$ denotes the dimension of input. The input  $\vec{h}_i^{spt}$ contains the attributes of the road link $v_i$ introduced in the Definition 1. We then apply a linear transformation to input $\vec{h}_i^{spt}$ using a learnable weight matrix $W \in \mathbb{R}^{F' \times F}$, where $F'$ is the dimensions of the transformed feature space. This step is pivotal for projecting the features into a new space where the spatial relationships can be more effectively analyzed. 

\paragraph{Attention Coefficients Calculation} Attention mechanisms are at the core of GATs, enabling the model to focus on the most relevant parts of the input graph. For each node $i$, the attention coefficients between itself and its directly connected neighborhood nodes $j \in \mathcal{N}_i$ are calculated by {\color{red} the following equation},
\begin{linenomath}
\begin{equation}\label{eq:6}
e_{ij} = \text{LeakyReLU}\left(a\left(W\vec{h}_i^{spt} \Vert W\vec{h}_j^{spt}\right)\right)
\end{equation}
\end{linenomath}
where $a(\cdot)$ is a shared attentional mechanism that signifies the importance of node $j$'s features to node $i$, $a: \mathbb{R}^{F} \times \mathbb{R}^{F'} \rightarrow \mathbb{R}$, and $(\Vert)$ is the concatenation operation. In this study, based on \cite{velivckovic2017graph}, the attention mechanism $a(\cdot)$ is a single-layer feedforward neural network, parametrized by a weight vector $\vec{a} \in \mathbb{R}^{2F'}$ {\color{red} and followed by a LeakyReLU activation with a fixed negative-slope coefficient of 0.2.} The $e_{ij}$ denotes the computed attention coefficient. 
These raw coefficients are then normalized across all neighbors using the softmax function, as shown {\color{red}below},
\begin{linenomath}
\begin{equation}\label{eq:7}
\alpha_{ij} = \text{Softmax}_j(e_{ij}) = \frac{\exp(e_{ij})}{\sum_{k \in \mathcal{N}_i} \exp(e_{ik})}
\end{equation}
\end{linenomath}
This normalization step ensures that the attention coefficients are comparable across different nodes and neighborhoods.

Finally, the model aggregates the features of neighboring nodes (including itself) weighted by the computed attention coefficients to update the representation of each node,
\begin{equation}\label{eq:8}
(\vec{h}_i^{spt})' = \sigma\left(\sum_{j \in \mathcal{N}_i} \alpha_{ij} W\vec{h}_j^{spt}\right)
\end{equation}
where $(\vec{h}_i^{spt})'$ represents the updated feature at node $i$, $\sigma$ denotes a non-linear activation. We use ReLU as the activation function.

\paragraph{Multi-head Attention Calculation}
We adopt multi-head attention instead of single-head attention since it allows the model to not only stabilize the learning process but also to explore information from different representations \cite{velivckovic2017graph,vaswani2017attention}. {\color{red}In multi-head attention, the attention mechanisms are independently performed in $K_{h}$ heads, where $K_{h}$ denotes the number of independent attention units.} Each of these heads potentially focuses on different relationships or features within the graph, enhancing the model's ability to capture complex patterns.
The outputs from these multiple heads are then averaged to produce the final output, which is expressed as, 

\begin{equation}
(\vec{h}_i^{spt})' = \sigma\left( \frac{1}{K_{h}} \sum_{k=1}^{K_h} \sum_{j \in \mathcal{N}_i} \alpha_{ij}^{(k)} W^{(k)} \vec{h}_j^{spt} \right)
\end{equation}
\noindent
where \( \alpha_{ij}^{(k)} \) and \( W^{(k)} \) are the attention coefficients and weight matrices for the \( k \)-th head, respectively, and \( K_h \) is the number of heads. In this study, we choose \( K_h = 2 \) to balance model complexity and computational efficiency, as two heads are sufficient to capture the necessary diversity in spatial representations without introducing excessive computational cost.

\begin{figure}[t]
\centering
  \includegraphics[width=0.48\textwidth]{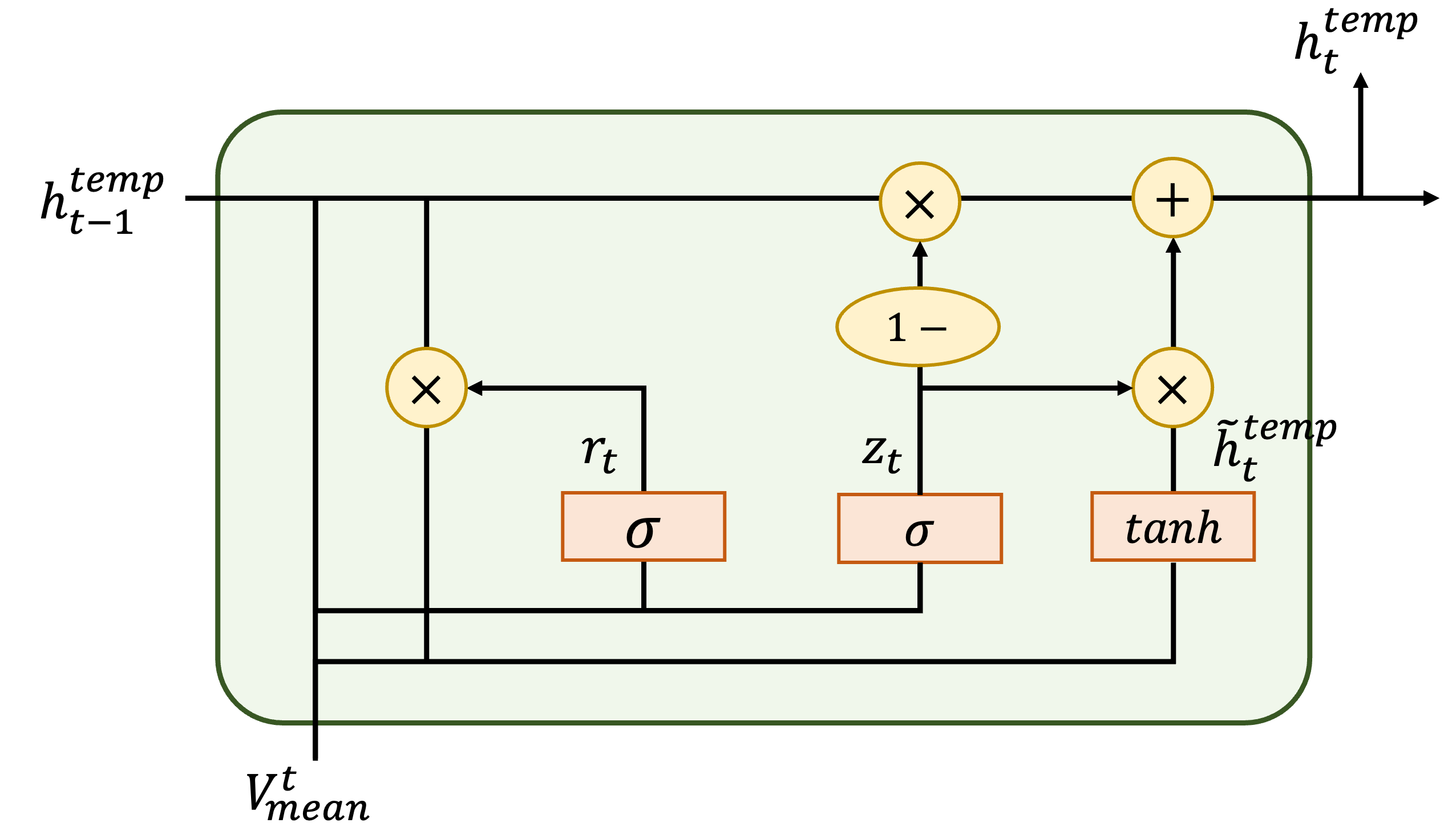}
  \caption{The model architecture of GRU}\label{fig:GRU}
\end{figure}
\noindent

\subsubsection{Temporal Embedding}
In the concept of the proposed LCF, we aim to estimate the speed of each link at time $t$, leveraging the network mean speed and the network configuration. To enhance the estimation accuracy, our research incorporates an analysis of temporal data, specifically the network mean speed from previous time steps, ranging from $t-t'$ to $t$. This addition not only enhances the model's ability to estimate speed across the network but also takes into consideration the temporal correlations associated with congestion dynamics. 
Therefore, the LCF can be re-defined as {\color{red}follows},
\noindent
\begin{linenomath}
\begin{equation}\label{eq:9}
\text{LCF}^{t} = f_{\text{LCF}}(G; [V_{\text{mean}}^{t-t'},V_{\text{mean}}^{t-t'+1}, \ldots, V_{\text{mean}}^{t}])
\end{equation}
\end{linenomath}

For the effective extraction of this sequential information, this study employs {\color{red}GRUs}. The models are widely used for processing time series data, capturing relevant historical context while mitigating common challenges such as the vanishing gradient problem \cite{cho2014properties}. Compared to LSTM, GRU has a simpler structure with faster convergence speed, higher computational efficiency, and fewer parameters for training. The model architecture, depicted in Fig.~\ref{fig:GRU}, revolves around the GRU's capability to update its hidden state $\vec{h}_t^{temp}$ at each timestep $t$. This update depends crucially on both the current input $V^t_{\text{mean}}$ and the previous hidden state $\vec{h}_{t-1}^{temp}$. The mathematical formulation of the GRU's update process is as follows,
\noindent
\begin{linenomath}
\begin{align}
z_t &= \sigma(W_z \cdot [\vec{h}_{t-1}^{temp}, V^t_{\text{mean}}] + b_z) \\
r_t &= \sigma(W_r \cdot [h_{t-1}^{temp}, V^t_{\text{mean}}] + b_r) \\
\tilde{h}_t^{temp} &= \tanh(W \cdot [r_t \ast \vec{h}_{t-1}^{temp}, V^t_{\text{mean}}] + b) \\
\vec{h}_t^{temp} &= (1 - z_t) \ast \vec{h}_{t-1}^{temp} + z_t \ast \tilde{h}_t^{temp}
\end{align}    
\end{linenomath}

where $z_t$ and $r_t$, represent the update and reset gates, respectively, controlling the extent of past information to be carried forward. In addition, $\tilde{h}_t^{temp}$ represents the candidate activation state. The symbols $W_z$, $W_r$, and $W$ denote the learnable weight matrices, $b_z$, $b_r$, and $b$ are the corresponding bias vectors. We use the sigmoid function $\sigma$ as the activation function. Through this structure, the GRU discerns and retains essential information in time steps, significantly improving the capacity of the proposed hybrid LCF model to accurately estimate the speed of each link by incorporating temporal information.


Finally, we obtain spatio-temporal embedded data that combines the outputs from spatial and temporal embeddings. By incorporating comprehensive spatial features, including dynamic bus lane configurations and leveraging the {\color{red}GATs} to account for network topology, along with temporal sequences of network mean speeds, we ensure that the spatio-temporal embeddings uniquely represent each scenario. This new input data is then processed through fully connected layers to estimate the speed of each link at time $t$. These layers are particularly adept at this task due to their ability to analyze complex patterns within the input data. Further discussions on the deep learning model configurations, parameter settings, and input shapes are provided in the later section.

\subsection{Traffic Simulation Model}

Dynamic traffic simulation is performed based on the macroscopic, queue-based Store-and-Forward (SaF) paradigm (see \cite{aboudolas2009store}), with enhanced structural properties derived by S-Model (see \cite{lin2011fast}). This is a link-based model representing vehicular traffic in a form of a fluid, which moves in a system of interconnected, controlled tubes of limited storage capacity and saturation flow. It was preferred over detailed car-based microscopic models due to its significantly lower computational cost, which was of paramount importance in view of the numerous simulation runs that would be necessary in the scope of this work. Due to its macroscopic nature, the information stored within the simulation is link-based (each link monitors a vehicle queue), which allowed for the numerous simulation runs of Barcelona city network (for different travel demand and network configurations) required for the training of the model to be executed in reasonable time. Moreover, the dynamic characteristics of congestion propagation (queue spill-backs) are successfully taken into consideration by the chosen model, thus providing the required modeling accuracy in estimating travel time, queue formation and proper MFD estimation. The details and the mathematical structure of the aforementioned model can be found in \cite{tsitsokas2021modeling}, therefore detailed model description is omitted. Instead, a brief qualitative description of the model's properties is given below. 

The traffic model operates on a network represented in the form of a directed graph $(N, Z)$\footnote{In this work, $(N, Z)$ represents the real physical network in the traffic simulator, where $N$ denotes intersections (nodes) and $Z$ denotes road links (edges). This is distinct from $G = (\mathcal{V}, E)$ defined in Section II-A, where $\mathcal{V}$ corresponds to physical network links used as model input.} and replicates traffic flow by updating the number of vehicles (or queue) inside every link $z$, $x_z(k)$, according to a time-discretized flow conservation equation, where the queue change rate is equal to the difference of inflow and outflow. Vehicle inflow is generated based on a dynamic Origin-Destination demand matrix, while routing is determined by turn ratios. Link inflows are equal to the sum of incoming flows from all upstream links plus the newly generated demand. Outflows are equal to outgoing flows to all downstream links and trip endings inside the link. Backward propagation of congestion is properly modeled by utilizing a transit flow calculation method that takes into account space availability downstream. More specifically, the model imposes zero transfer flow for the following time step if the receiving link is already congested (the queue at the current time step is close to the link's storage capacity). Therefore, in case of increased traffic, queues propagate backward. 

The inherent inaccuracy in estimating travel time and link queue length, owed to the dimensionless nature of the original SaF version has been minimized by integrating the structure of S-Model \cite{lin2011fast}, where every link queue $x_z$ consists of two distinct queues, $m_z$ and $w_z$, representing vehicles moving and waiting at the link end, respectively. Transit flow vehicles that enter a new link first join the moving part of the link, where they are assumed to move with free-flow speed. They transfer to the queuing part after a number of time steps determined by the actual position of the queue end. A binary function dictates whether an approach of an intersection gets green light at the current time-step, according to the active signal plan. If an approach gets red light, this function will set the current outflow to zero for this time step, regardless of the state of the queues upstream or downstream. 

The mass conservation equation is transformed into a recursive formula that calculates the link queues in every time step based on the queues of the previous time step and the respective outflows of the current time step. By applying these formulas for a predefined number of time steps, we get a complete traffic simulation for a specified demand scenario. By knowing the queues of every network link at every time step (state variables $m_z$ and $w_z$, $\forall z \in Z$), we can compute the total travel time of all vehicles. Mean link speed is estimated for every network link $z \in Z$ based on link outflow and queue values, according to the following relation,
\begin{equation}\label{eq:v_car}
    v_z(t) = \mbox{min} \left( v_{\textrm{ff}},  \frac{\sum_{j \in T_w(t-1)}{u_z(j)} L_z }{\sum_{j \in T_w(t-1)}{x_z(j)}} \right) \geq v_{\min}  
\end{equation}
where $t$ is the time window index, $v_{\textrm{ff}}$ is the free flow speed, $T_w(t)$ is the set of simulation time step indices corresponding to time window $t$, $L_z$ is the length of link $z$, and $u_z(j)$ and $x_z(j)$ denote the outflow {\color{red}(veh/hour)} and accumulation {\color{red} (i.e. the total number of vehicles present in the the link $z$ }at simulation time step $j$, respectively.  

In this study, we use fixed turning ratios derived from a distance-based shorted path algorithm applied to our OD matrix. While more advanced routing strategies could have been applied, exploring them falls outside the scope of our research objectives, and we prefer not to add complexity to the data generation.

\label{sec:methodology}

\section{Performance Evaluation}
\subsection{Data Description}

\begin{figure}[!t]
  \centering
  \includegraphics[width=0.45\textwidth]{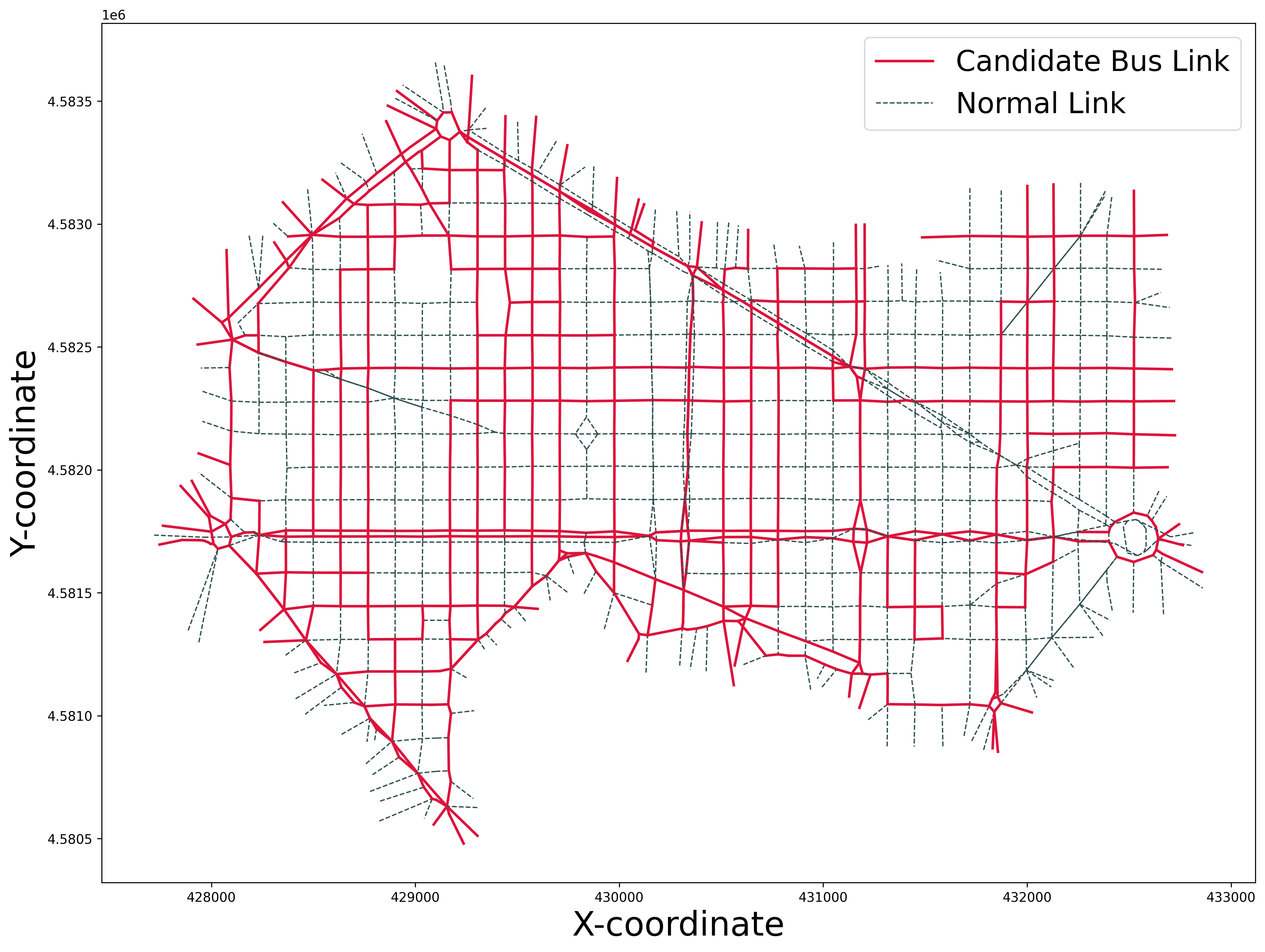}
  \caption{Road network of Barcelona with candidate bus links}
  \label{fig: bus network}
\end{figure}

The proposed method is applied to the city center of Barcelona, Spain, illustrated in Fig.~\ref{fig: bus network}. The network comprises 933 nodes and 1570 road links, with 565 signalized junctions operating on fixed-length signal control cycles ranging from 90 to 100 seconds. The road links vary from 1 to 5 lanes. Out of all the links, 818 labeled in red are candidates for the bus lane setting. These are in principle all links that are included in the route of at least one bus line and have two or more lanes in total. The average free-flow speed for both cars and buses is assumed to be 25 km/h. The travel demand scenarios for the simulation experiments consist of a 15-minute warm-up period, followed by a 1.75-hour constant peak demand phase, culminating in a total simulation duration of 6 hours to simulate the morning traffic peak. The considered trip volume for each Origin-Destination (OD) is based on real traffic data from the city of Barcelona in terms of traffic distribution within the network, reflecting the trips to and from the city center. However, given the constraints that usually come with simulation of different scenarios in terms of emptying the network at the end, the total number of trips of the original demand was transformed into a 1.75 hours peak-demand, following a 15 minutes interval of lower ``warm-up" demand. In the interval between 2 to 6 hours of simulation, no new demand was considered, in order to allow sufficient time for the network to release all the vehicles. We aggregate the simulation data at 3-minute intervals, dividing each simulation scenario into 120 steps. Since this study focuses exclusively on the 2-hour loading phase, we select only the first 40 steps for each scenario. Thus, our objective is to estimate the speed of each link at each of these steps, corresponding to every 3 minutes during the loading phase.
\FloatBarrier
To enhance the robustness of our speed estimation model, we generate a diverse set of experiment data reflecting varying lane configurations, OD distributions, and total demands. The variability in terms of lane configuration is achieved by changing the assignment of dedicated bus lanes, which results in different link storage capacities and saturation flow. We focus on dedicated bus lanes since by assuming a different spatial allocation for them, we are able to study the impact of localized infrastructure variations on traffic flow, without fundamentally disrupting the network's overall topology. This approach provides a practical and controlled method to simulate different road network scenarios and understand how specific local modifications may affect link speeds. In terms of OD distribution, we alter the traffic volumes for each OD pair within a fluctuation range of ±20\%,  while the aggregate traffic volume remains constant. This approach mimics the day-to-day shifts in traffic distribution across the network, providing a more dynamic and realistic training environment for our model. After adjusting the OD distribution, we apply a random scaling factor to modify the total demand in each simulation. The magnitude of this scaling factor determines the demand level that each simulation represents—specifically, ``low'', ``medium'', or ``high'' demand levels. By varying the scaling factor in each simulation, we can simulate various traffic conditions, ranging from off-peak hours (low demand) to peak hours (high demand). This method allows us to assess the model's performance across a spectrum of traffic volumes and ensures its ability to adapt to changing traffic conditions.

\begin{figure*}[t]
\centering
\captionsetup[subfloat]{font=scriptsize} 
\subfloat[]{\includegraphics[width=0.24\linewidth]{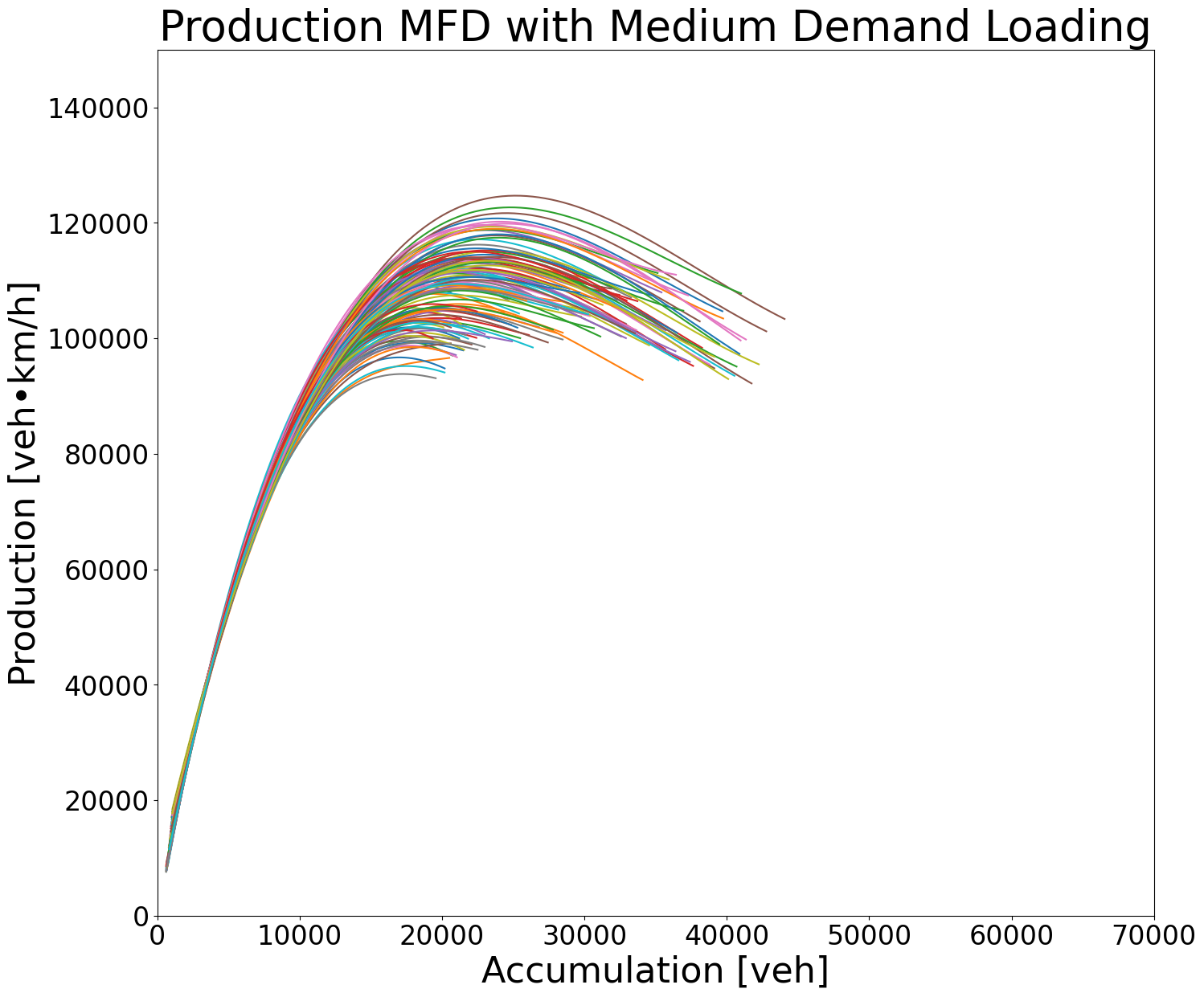}\label{fig:MFD_train}}
\hfil
\subfloat[]{\includegraphics[width=0.24\linewidth]{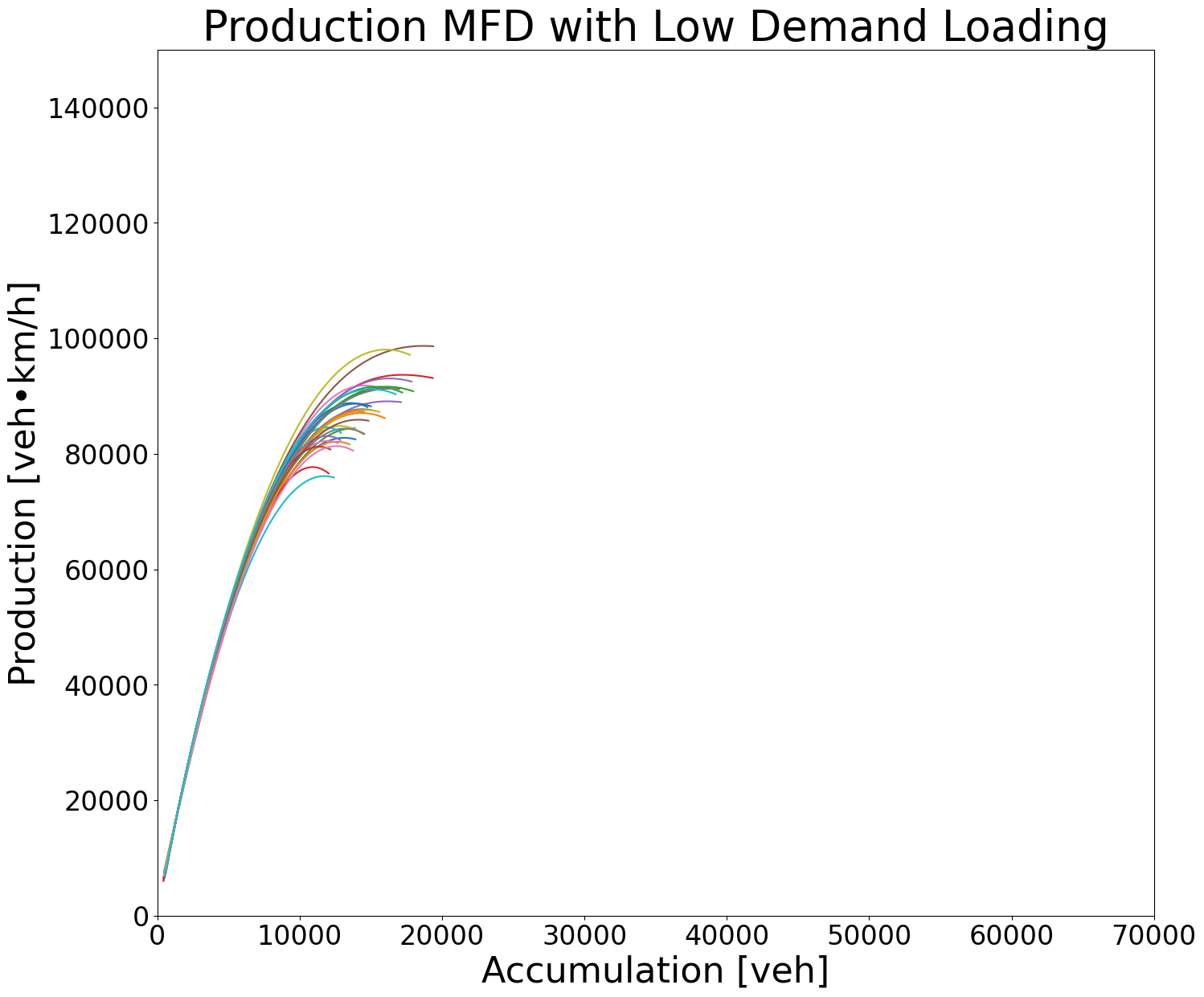}\label{fig:MFD_test_low}}
\hfil
\subfloat[]{\includegraphics[width=0.24\linewidth]{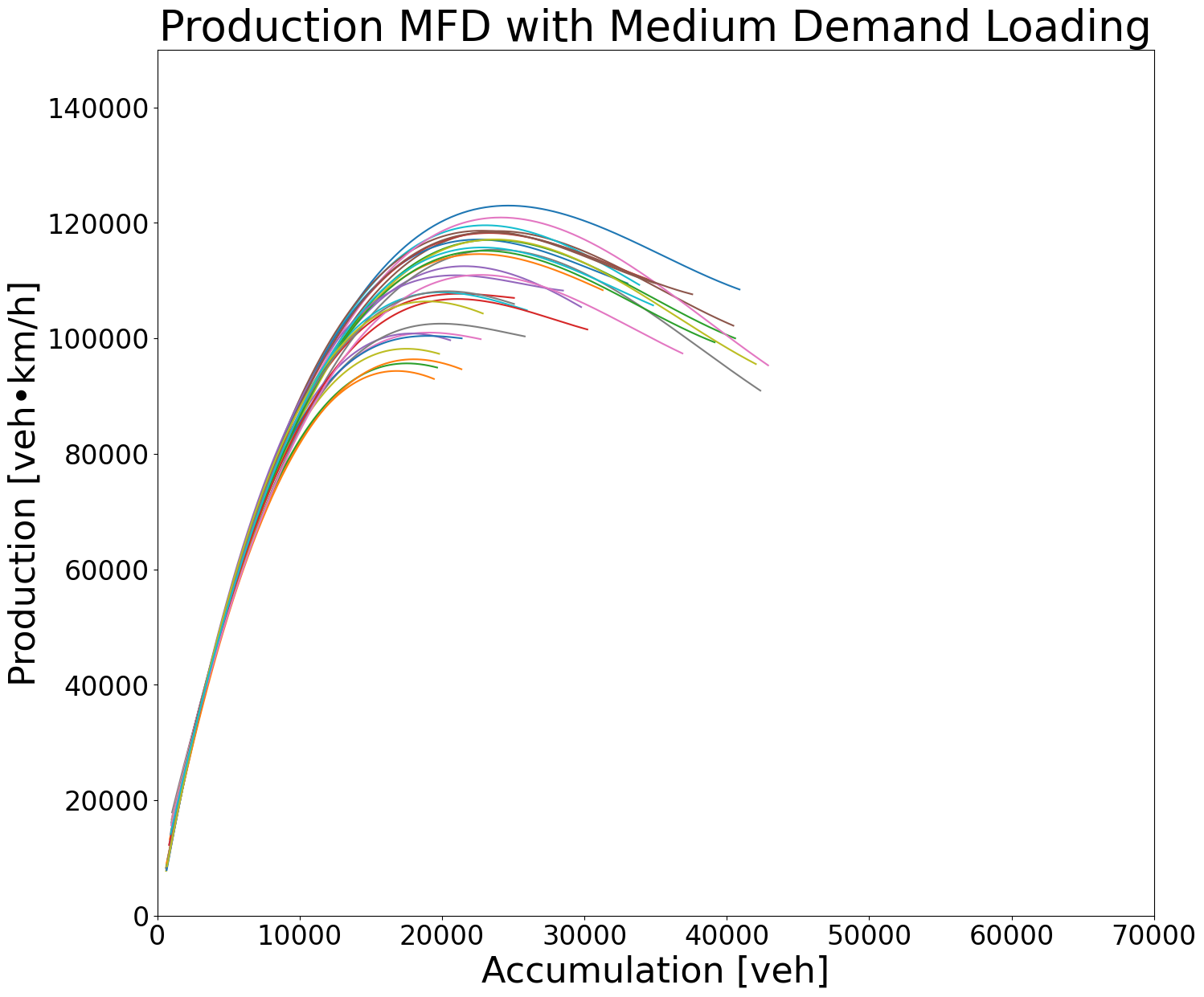}\label{fig:MFD_test_med}}
\hfil
\subfloat[]{\includegraphics[width=0.24\linewidth]{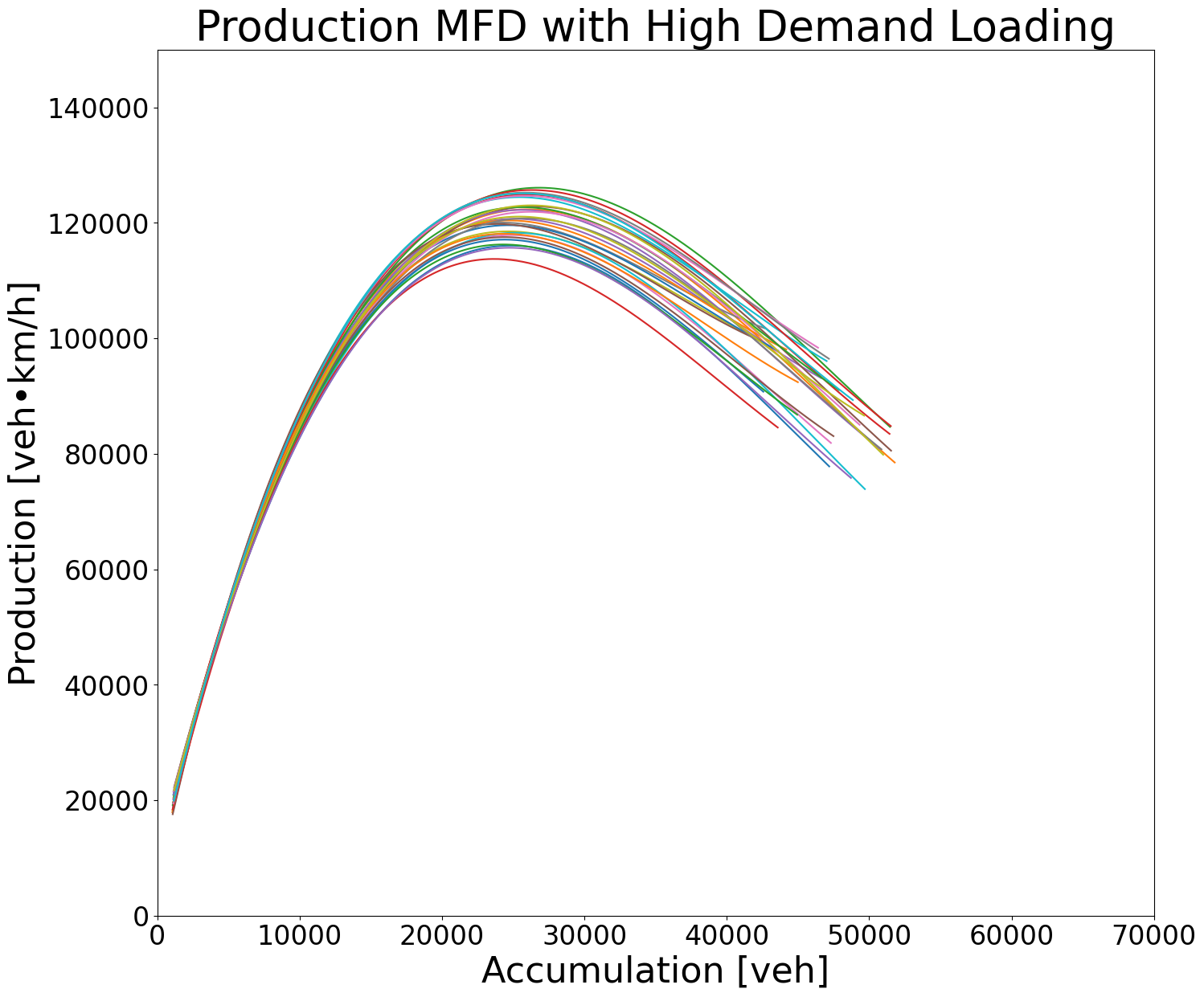}\label{fig:MFD_test_high}}
\caption{Production MFD of training and testing data: (a) training data with medium demand (b) testing data with low demand (c) testing data with medium demand (d) testing data with high demand. }
\label{fig:demand levels}
\end{figure*}

In generating the training data, we constrain the total loading demand to a medium level, representing moderately congested traffic conditions commonly observed in urban environments. Importantly, even within this medium-demand level, the total demand in each simulation varies within a specific range, thereby providing diverse traffic conditions for the model to learn from. We consider only the loading phase in our analysis to avoid the hysteresis loop that appears during network unloading. We fit the MFD functions during loading using regression techniques. Figure~\ref{fig:demand levels} (a) illustrates the diverse network conditions across various scenarios in the training data. Each curve in the figure represents a distinct scenario characterized by a unique OD matrix and network configuration, including variations in total demand, OD distributions and the allocation of dedicated bus lanes.
For the testing process, our strategy is to evaluate the model's performance under a variety of conditions. We evaluate the model's performance in three distinct cases: the first maintains the medium-demand level as in the training phase. The second and third cases explore scenarios with extreme demand levels—low and high—that also involve adjustments to total demand within their respective ranges and modifications to OD distributions and lane configurations. Figures~\ref{fig:demand levels} (b) to (d) illustrate the MFDs for these testing cases. In scenarios with the low-demand level, maximum production points are absent; conversely, congestion becomes evident in high-demand situations. This comprehensive testing approach is designed to evaluate the model's capacity to adapt and maintain accuracy across diverse and realistic traffic scenarios, thereby confirming its applicability to urban traffic management and planning. In our experiments, we utilize only a single calibrated speed MFD function derived from the entire training dataset. Our simulations indicate that loading patterns and network configurations do not significantly affect the MFD shape; therefore, a single mean speed curve is sufficient to fit all observations.

\subsection{Comparison Methods and Evaluation Metrics}

In this study, we compare our proposed speed estimation method with two popular machine learning models and three deep learning models to evaluate its effectiveness:

\begin{itemize}
    \item Linear Regression (LR)\cite{weisberg2005applied}: LR models a linear relationship between input features and link speeds using ordinary least squares.
    
    \item Extreme Gradient Boosting (XGBoost)\cite{chen2016xgboost}: XGBoost employs an ensemble of gradient-boosted decision trees to model complex relationships between inputs and outputs.
    
    \item Deep Neural Network (DNN)\cite{hornik1989multilayer}: DNNs utilizes multiple layers of non-linear transformations to capture intricate patterns in the data.
    
    \item {\color{red}GAT}\cite{velivckovic2017graph}: GATs leverage attention mechanisms to focus on the most relevant parts of the graph structure, capturing spatial dependencies within the network. See Section~\ref{subsec:method} for details. 
    
    \item Hybrid model of DNN and GRU (DNN-GRU): This hybrid model combines a DNN for processing spatial attributes and a GRU for capturing temporal dependencies through sequences of network mean speeds.
\end{itemize}

All baseline models utilize the spatial attributes of the road network \( G \) and the network mean speed \( V_{\text{mean}}^t \) derived from the speed MFD as inputs. Specifically, LR, XGBoost, DNN, and GAT rely on the current network mean speed \( V_{\text{mean}}^t \), while the DNN-GRU model additionally incorporates a sequence of network mean speeds from previous time steps, \( (V_{\text{mean}}^{t-t'}, V_{\text{mean}}^{t-t'+1}, \ldots, V_{\text{mean}}^{t}) \), to capture temporal dynamics.

For evaluating and comparing model performance, we use Mean Absolute Error (MAE), Root Mean Squared Error (RMSE), symmetric Mean Absolute Percentage Error (sMAPE), Mean of the Error (Err. Mean), and Standard Deviation of the Error (Err. STD) as our evaluation metrics. Each model is tested under the same conditions to ensure a fair comparison of their performance.

\subsection{Experiment Setup}
In this study, we construct a dataset composed of 150 randomly generated traffic simulations, divided into training, validation, and test subsets in a ratio of 7:1:2, respectively. {\color{red}Within each scenario, we collect the first 40 data samples—the network-loading period—with each sample representing the network’s state at a specific time step, capturing link speeds and other relevant variables.} This process leads to a total of 4,200 data samples for training, 600 for validation, and 1,200 for testing. During the model training phase, the input data for spatial embedding is organized into several batches randomly. Each batch included data for 1570 road links, with each link described by 10 distinct attributes. The total number of batches processed is 128, and the dimension of the hidden layers used for the spatial embedding is also set at 128. Temporal embedding is handled similarly, with each batch containing the same number of links, and the model considers an array of average speed for each link derived from the current and the four previous time steps, a total of five-time steps per link. If the data in the previous steps are insufficient or unavailable, a default padding value of -1 is used to maintain array consistency. The hidden dimension of the {\color{red}GRU} is also set to 128. The architecture of the speed estimator includes a fully connected layer comprising six layers. The dimensions of these layers change sequentially from 256 to 384, 256, 128, 64, 32, and finally down a single output unit. The AdamW optimizer is employed to minimize the loss function, with an initial learning rate of 0.002. The learning rate is adjusted using the stepLR scheduler, with a step size of 80 and a decay factor (gamma) of 0.85. Hyperparameters, including batch size, hidden layer dimensions, learning rate, and scheduler settings, were selected through meticulous manual tuning to achieve an optimal balance between model performance and computational cost. All experiments are conducted on a Ubuntu server (Intel(R) Xeon(R) Gold 5218 CPU @ 2.30GHz, GPU NVIDIA GeForce RTX 2080 Ti).

\subsection{Results} 

The results for the proposed speed estimation method are discussed in four perspectives: \textit{1) Network Partitioning Visualization, 2) LCF Performance Evaluation, 3) LCF Robustness Evaluation, and 4) LCF Applicability in Simulation}. Our analysis begins with a visualization of network partitioning. Then, we evaluate the performance of our proposed LCF with the medium-demand level but varying OD distributions and road configuration. The robustness of LCF is further evaluated against different demand levels — both low and high. Finally, we test our proposed LCF to estimate the travel time along with random paths. In the following sections, each of them will be discussed in detail.

\subsubsection{\textbf{Network Partitioning Visualization}}

\begin{figure}[t]
\centering
  \includegraphics[width=0.45\textwidth]{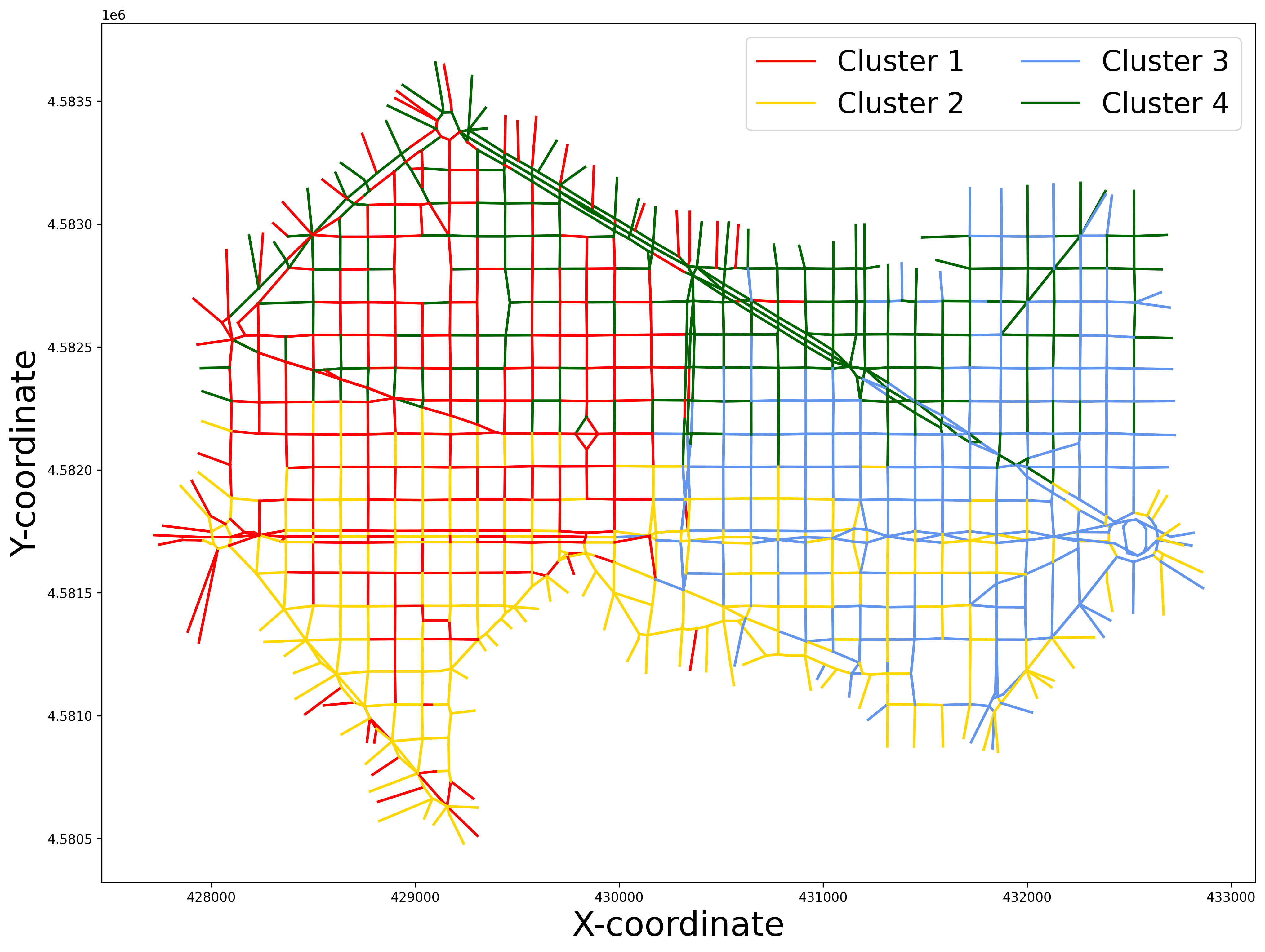}
  \caption{Visualization result of network partitioning with 4 clusters}\label{fig: network visualization}
\end{figure}

Fig.~\ref{fig: network visualization} presents the visualization result of network partitioning. In this study, we divide the road network into 4 sub-regions, where clusters 1 to 4 are associated with 369, 404, 359, and 438 links, respectively. The proposed partitioning method leads to a deliberate symmetry structure, where clusters 1 (red) and 3 (blue), as well as clusters 2 (yellow) and 4 (green), are positioned symmetrically across the network. Additionally, each pair of clusters contains a similar number of links. This symmetry characteristic arises from the consideration of each link's location, ensuring the importance of spatial configuration in the clustering process. 

However, the boundaries of these clusters are not distinctly defined due to the integration of speed data into the clustering process. Specifically, clusters 2 (yellow) and 4 (green) show a spread-out pattern, which mainly indicates the boundary area with varied traffic conditions. Conversely, clusters 1 (red) and 3 (blue) display a more grid-like configuration, suggesting areas with more uniform road structures with more consistent traffic conditions. 

This clustering strategy, considering both the geographical location of each link and the mean speed of the network, aims to mathematically enhance the speed estimation performance. The visualized results suggest that the method is capable of considering the global and local characteristics of the network, potentially facilitating the development of a high-performing LCF. While compact cluster shape might be important as shown in the literature for perimeter control applications, it does not influence our estimation results, as our ultimate goal is to obtain accurate link correction factors. 

\noindent

\subsubsection{\textbf{LCF Performance Evaluation}}

We evaluate our proposed speed estimation approach, which utilizes the {\color{red}LCF} calculated by a deep learning function incorporating GATs and GRUs, against six baseline models using the test dataset. This test dataset maintains a consistent medium-demand level with the training data but varies in ODs and road configurations. The test results, which compare the estimated link speeds with the actual simulated link speeds, are presented in Table~\ref{tab:1}. Models enhanced with network partitioning information are denoted with ``-P", highlighting the impact of this feature on performance.

\begin{table}[t]
\setlength{\arrayrulewidth}{0.25pt}
\centering
\caption{PERFORMANCE EVALUATION OF LCF AT LINK LEVEL [KM/H]}
\label{tab:1}
\renewcommand{\arraystretch}{1.3} 
\resizebox{0.45\textwidth}{!}{%
\begin{tabular}{c|ccccc}
\hline
\noalign{\vskip-0.5pt} 
\hline
Model                   & MAE    & RMSE   & sMAPE(\%) & Err.Mean & Err.STD \\ \hline
MFD                     & 5.67   & 6.80   & 61.78     & 0.00     & 6.80    \\
MFD-P                   & 4.37   & 5.43   & 51.61     & 0.00     & 5.43    \\
LR                      & 5.30   & 6.43   & 58.62     & 0.01     & 6.43    \\
LR-P                    & 4.24   & 5.28   & 50.77     & 0.02     & 5.28    \\
XGBOOST                 & 2.91   & 3.77   & 40.21     & 0.13     & 3.77    \\
XGBOOST-P               & 2.56   & 3.37   & 36.01     & 0.12     & 3.37    \\
DNN                     & 2.21   & 3.85   & 29.29     & 0.09     & 3.85    \\
DNN-P                   & 1.79   & 3.25   & 25.22     & 0.04     & 3.25    \\
GAT                     & 1.44   & 2.86   & 21.56     & 0.03     & 2.85    \\
GAT-P                   & 1.32   & 2.67   & 20.23     & 0.05     & 2.67    \\
DNN-GRU                 & 1.35   & 2.69   & 20.60     & 0.10     & 2.69    \\
DNN-GRU-P               & 0.84   & 1.92   & 14.45     & -0.07    & 1.92    \\
GAT-GRU                 & 1.09   & 2.42   & 17.65     & -0.02    & 2.42    \\
GAT-GRU-P               & 0.76   & 1.82   & 13.42     & 0.01     & 1.82    \\ 
\hline
\noalign{\vskip-0.5pt} 
\hline
\end{tabular}%
}
\end{table}

\begin{figure}[t]
\centering
  \includegraphics[width=0.45\textwidth]{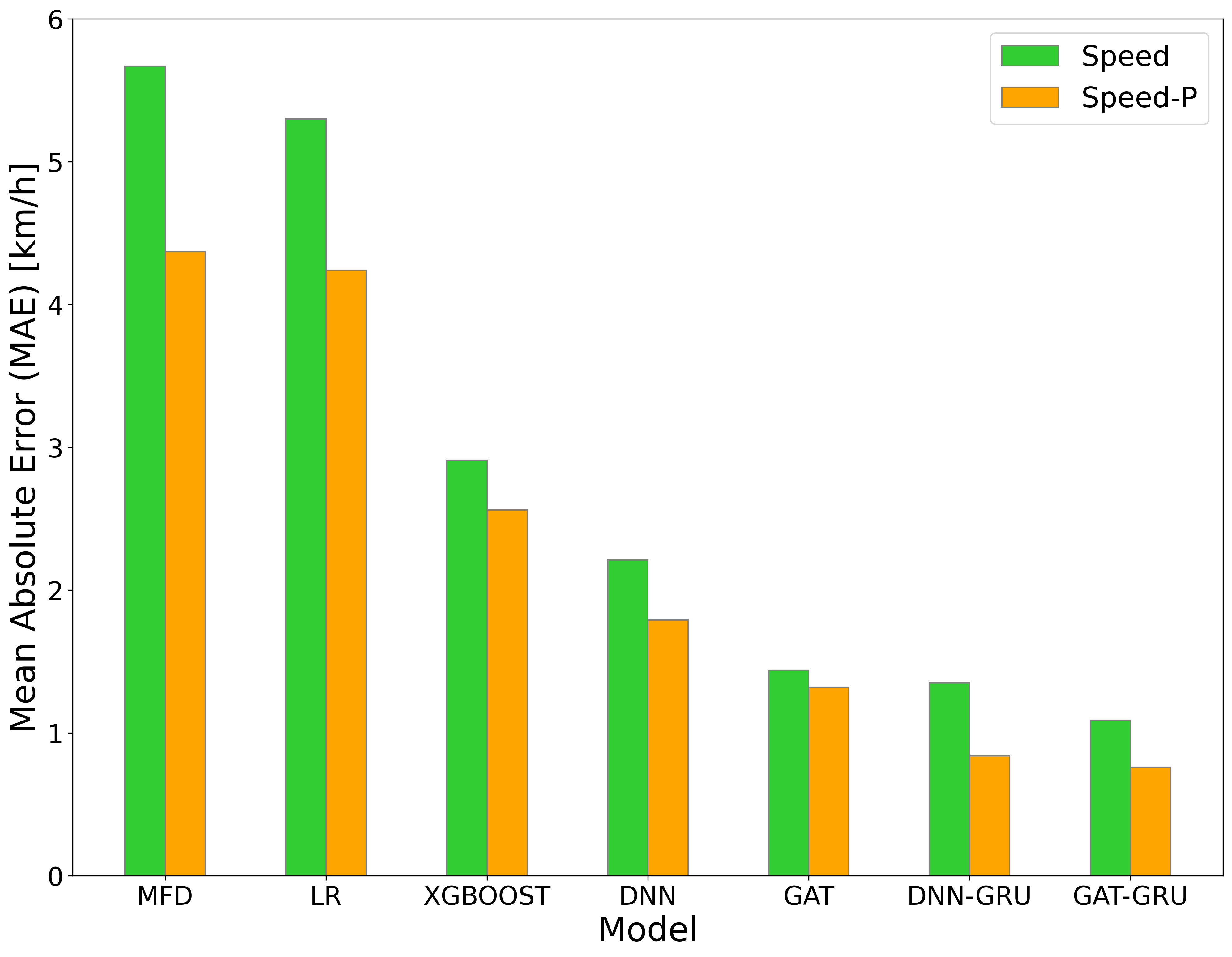}
  \caption{Performance comparison between with/without network partitioning (Speed-P indicates results of adding partitioning information)}\label{fig: NP statistics}
\end{figure}

\begin{figure}[t]
\centering
  \includegraphics[width=0.45\textwidth]{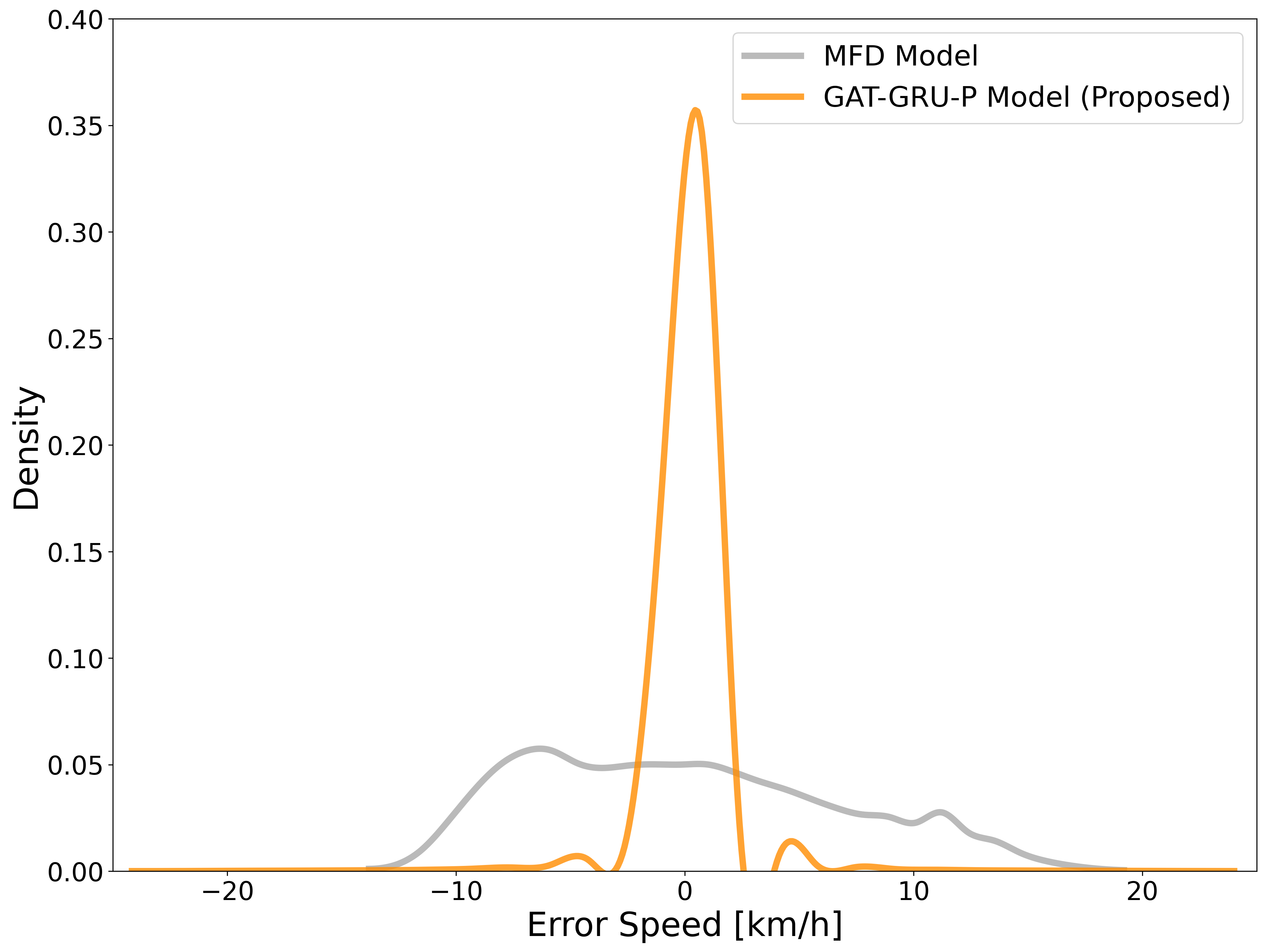}
  \caption{Comparison of speed estimation error distribution between MFD and proposed model}\label{fig: Error Speed Distribution}
\end{figure}
From the perspective of the models to calculate LCF, our analysis demonstrates the superiority of deep learning models over traditional machine learning approaches in estimating road network speed. Specifically, the Linear Regression (LR) model and its variant LR-P, show similar performance to the MFD model across three metrics. In addition, even though the XGBoost shows the ability to estimate the speed, the performance doesn't match the effectiveness of deep learning models. This indicates the limitation of traditional machine learning models in solving complex, non-linear problems that deep learning models can adeptly address. Among the deep learning models, our proposed GAT-GRU model, achieving an MAE of 1.09 km/h, an RMSE of 2.42 km/h, and an sMAPE of 17.65 \%, notably outperforms all baseline models. Specifically, compared to the MFD model, our model demonstrates improvements of approximately 80.8\% in MAE, 64.4\% in RMSE, and 71.4\% in sMAPE. Furthermore, the proposed GAT-GRU model shows enhancements over the GAT model, which shows the MAE of 1.44 km/h, RMSE of 2.86 km/h, and sMAPE of 21.56\% marking a 24.3\% reduction in MAE, a 15.4\% reduction in RMSE and 18.1\% reduction in sMAPE. These comparisons underscore the effectiveness of integrating spatial and temporal data, enhancing the model's estimation accuracy and establishing its superior performance across multiple baselines. Compared to the hybrid DNN-GRU model, the proposed model lowers the MAE, RMSE, and sMAPE by 19.3\%, 10.0\%, and 12.8\%, respectively. Therefore, the results indicate the effectiveness of the proposed GAT-GRU, which leverages GATs with GRUs for temporal embedding, in the speed estimation. 

The impact of network partitioning is another notable perspective of the analysis. {\color{red} All models enhanced with network partitioning information show significant improvements in speed estimation, as summarized in Table~\ref{tab:1} and visualized by the MAE comparison histogram in Fig.~\ref{fig: NP statistics}.} For example, the GAT-GRU model with partitioning (GAT-GRU-P) reduces the MAE to 0.76 km/h, the RMSE to 1.82 km/h, and the sMAPE to 13.42\% from the non-partitioned GAT-GRU's MAE of 1.09 km/h, RMSE of 2.42 km/h, and sMAPE of 17.65\%, marking improvements of approximately 30.3\% in MAE, 24.8\% in RMSE, and 24.0\% in sMAPE respectively. Across all models, the average improvement due to network partitioning is around 21.5\% in MAE, 17.7\% in RMSE, and approximately 16.3\% in sMAPE. These enhancements highlight the effectiveness of dividing the road network into more homogeneous sub-regions, which facilitates a more accurate capture of localized traffic patterns, thereby significantly enhancing the accuracy of traffic speed estimation. 

\begin{figure*}[!t]
\centering
\captionsetup[subfloat]{font=scriptsize} 
\subfloat[]{\includegraphics[width=0.24\linewidth]{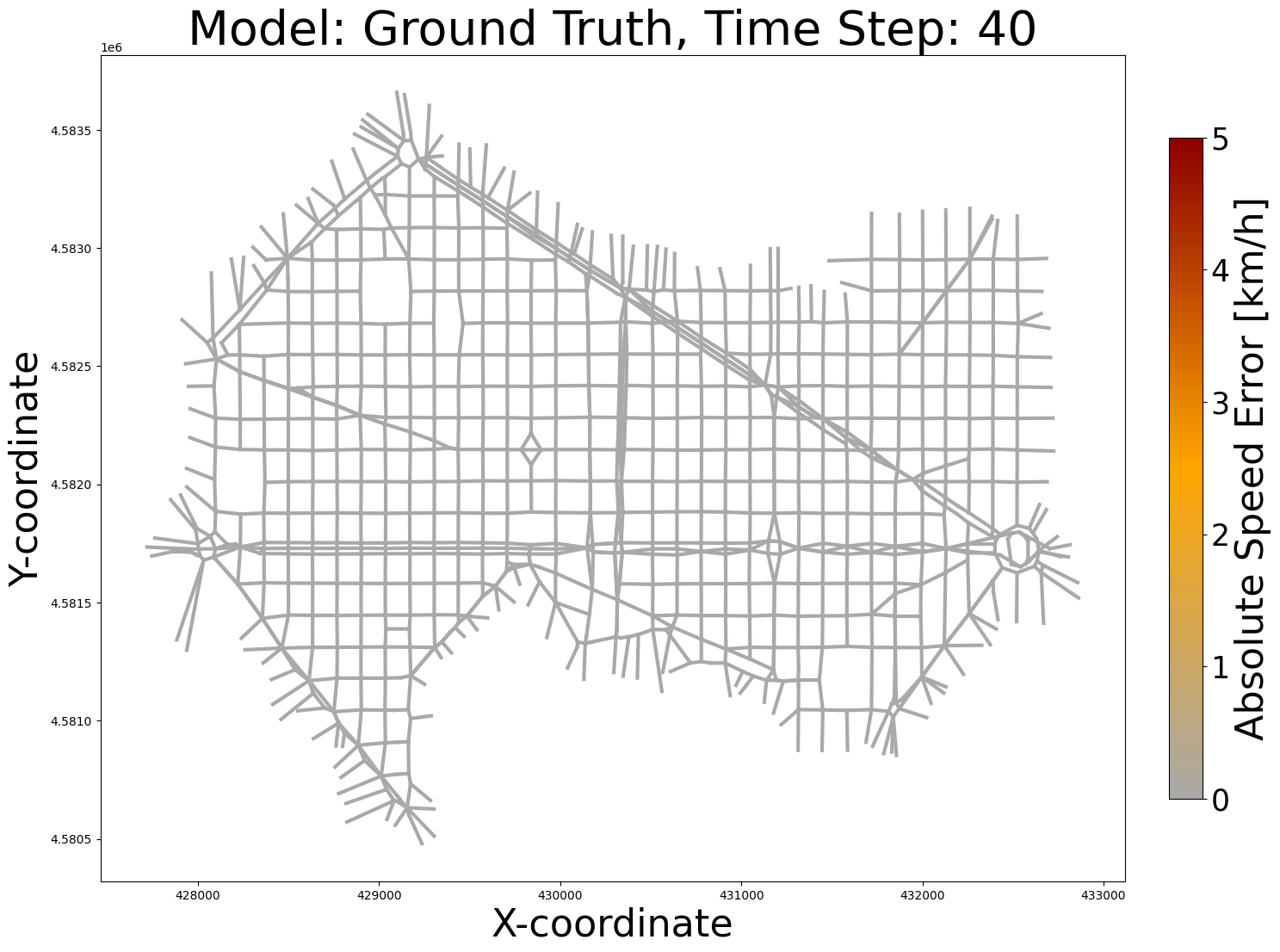}\label{fig:ground_truth_result}}
\hfil
\subfloat[]{\includegraphics[width=0.24\linewidth]{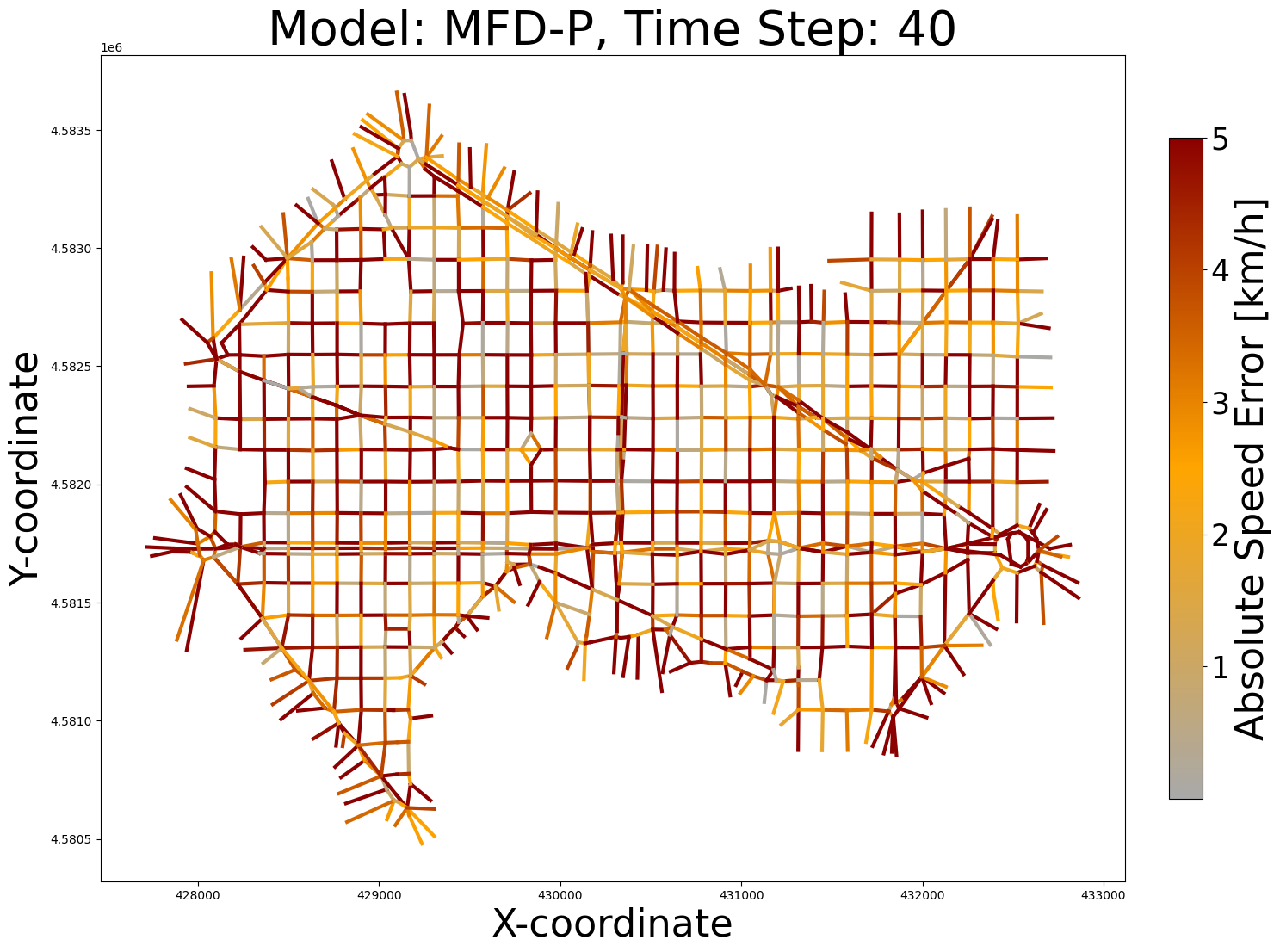}\label{fig:MFD_result}}
\hfil
\subfloat[]{\includegraphics[width=0.24\linewidth]{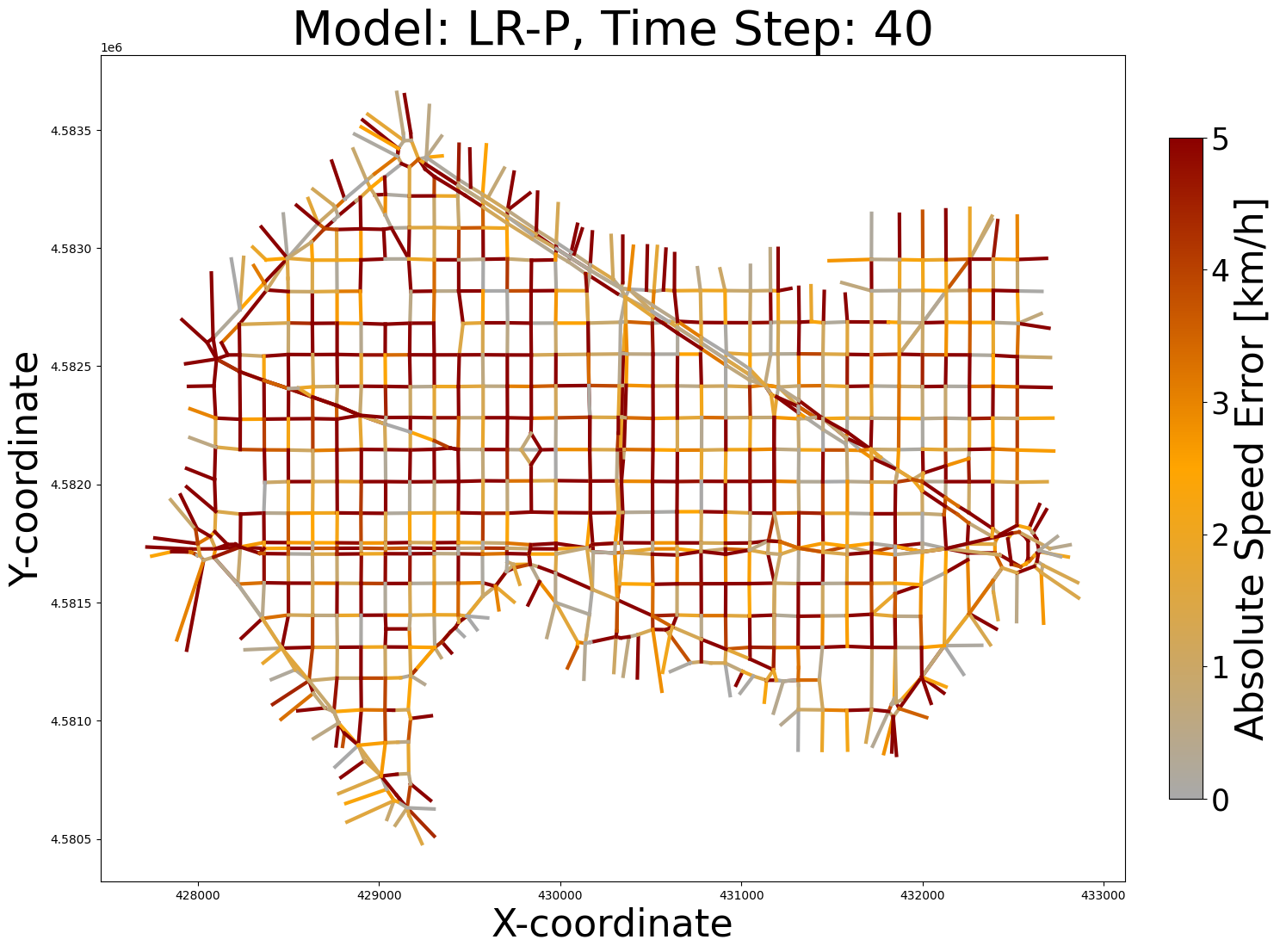}\label{fig:LR_result}}
\hfil
\subfloat[]{\includegraphics[width=0.24\linewidth]{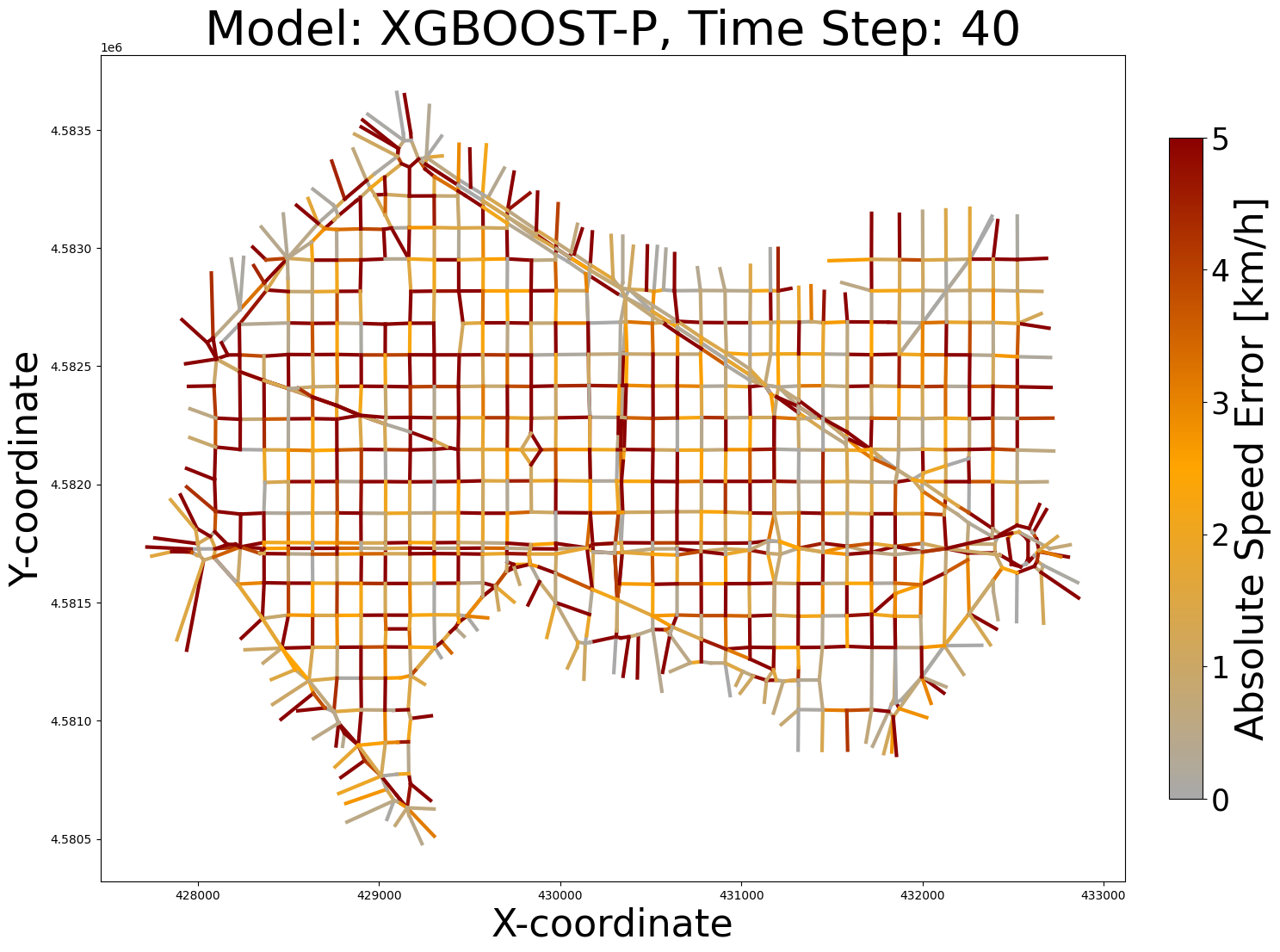}\label{fig:XGBOOST_result}}
\\ 
\subfloat[]{\includegraphics[width=0.24\linewidth]{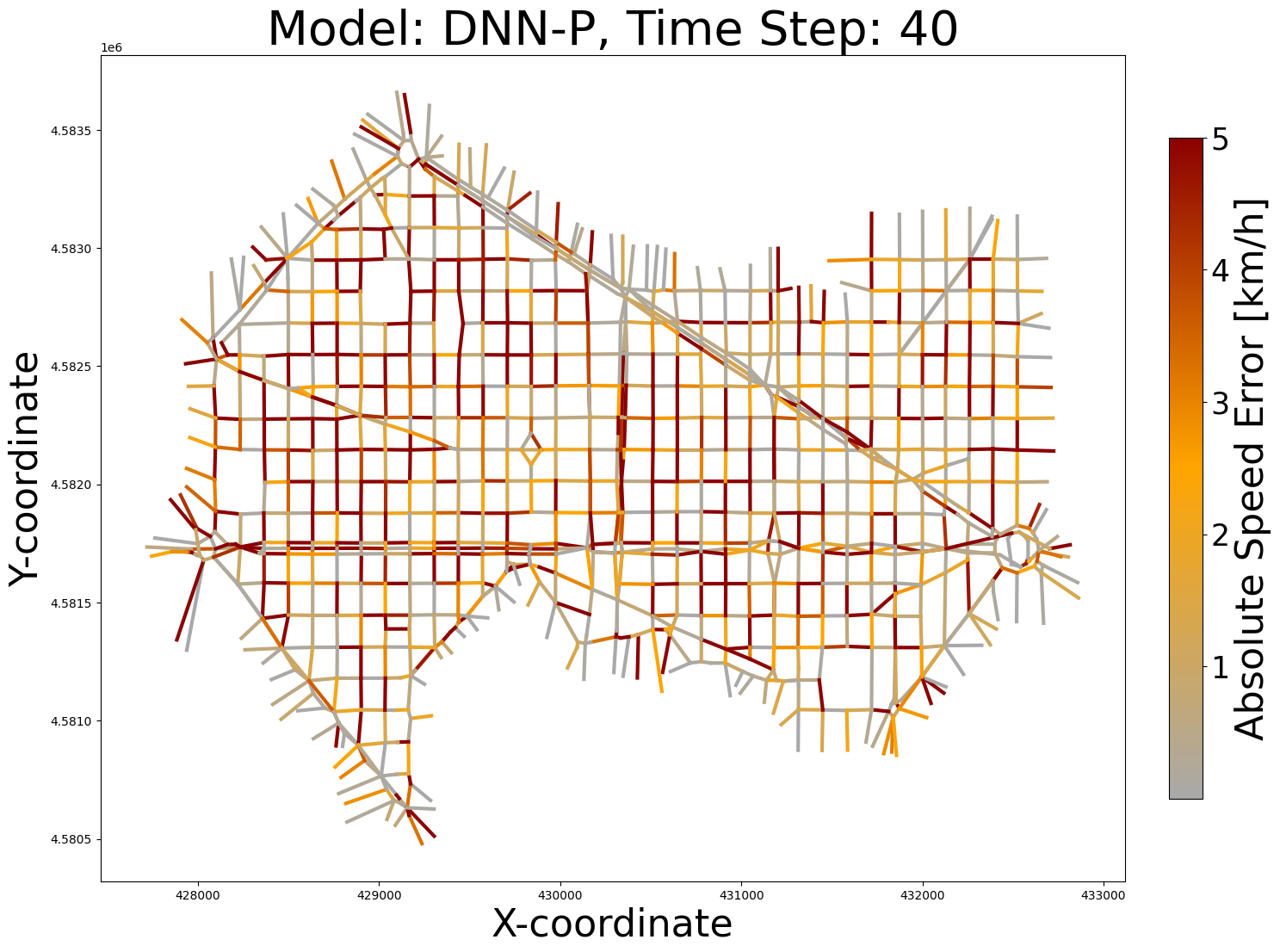}\label{fig:DNN_result}}
\hfil
\subfloat[]{\includegraphics[width=0.24\linewidth]{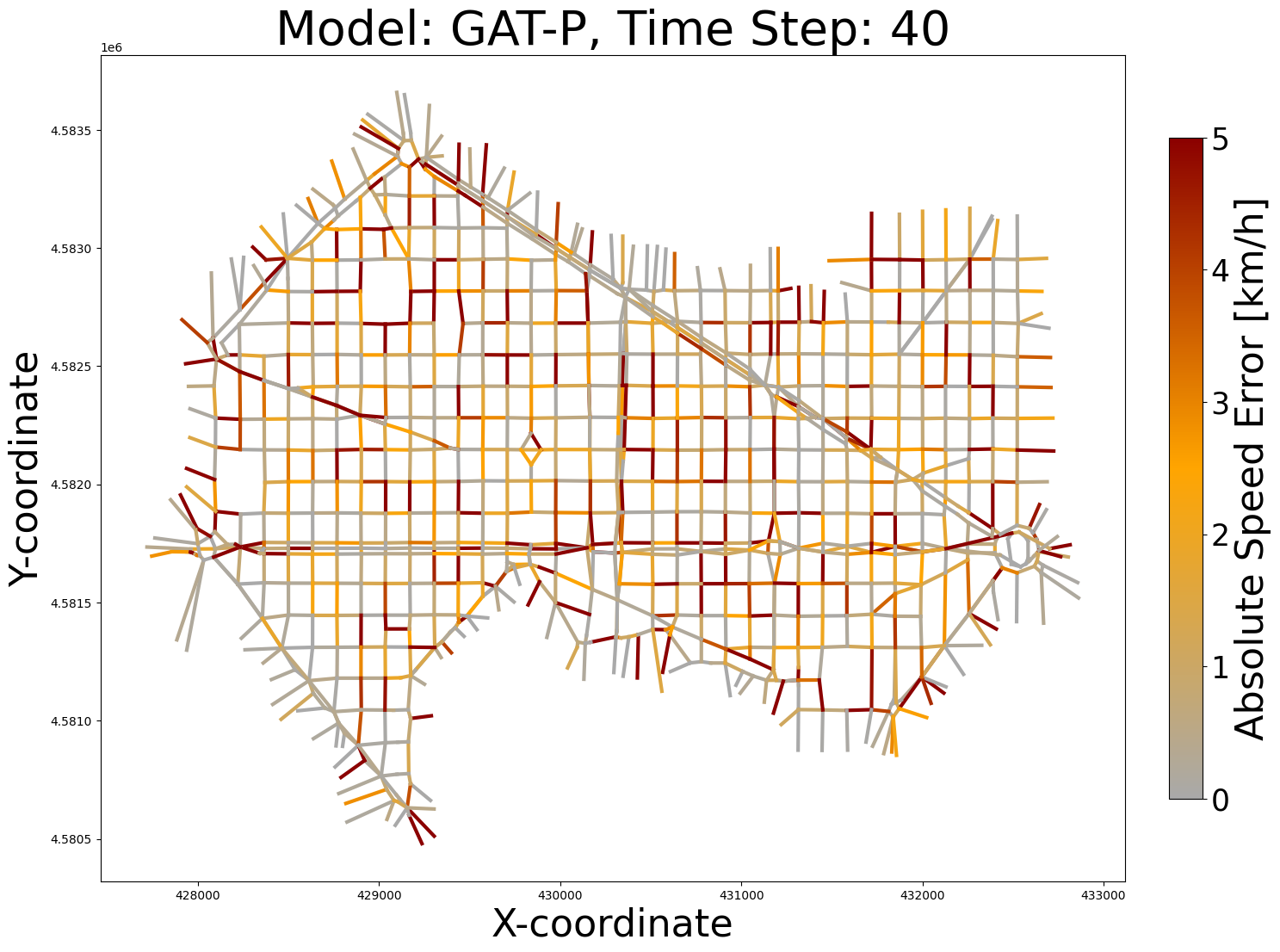}\label{fig:GAT_result}}
\hfil
\subfloat[]{\includegraphics[width=0.24\linewidth]{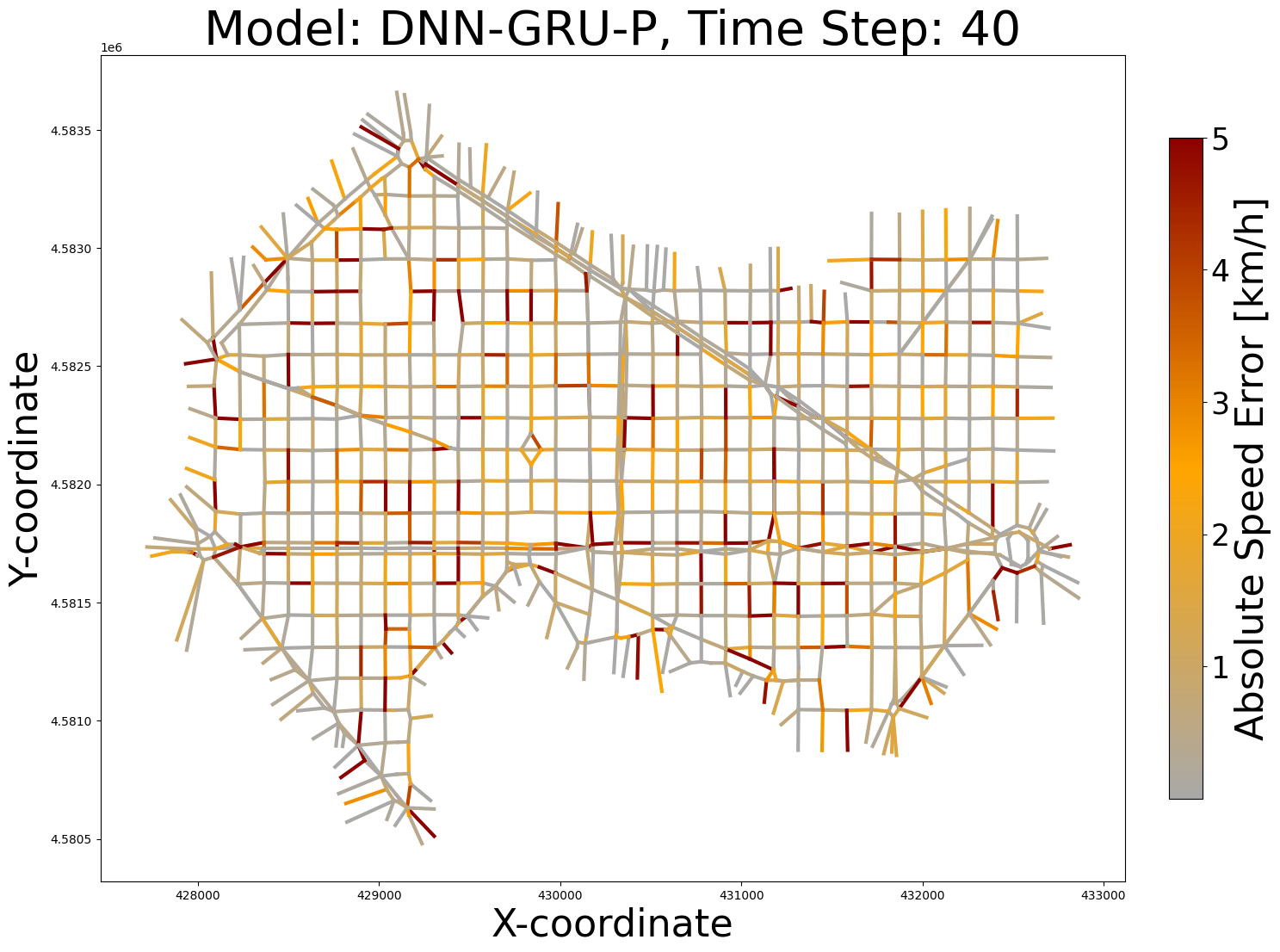}\label{fig:DNN-GRU_result}}
\hfil
\subfloat[]{\includegraphics[width=0.24\linewidth]{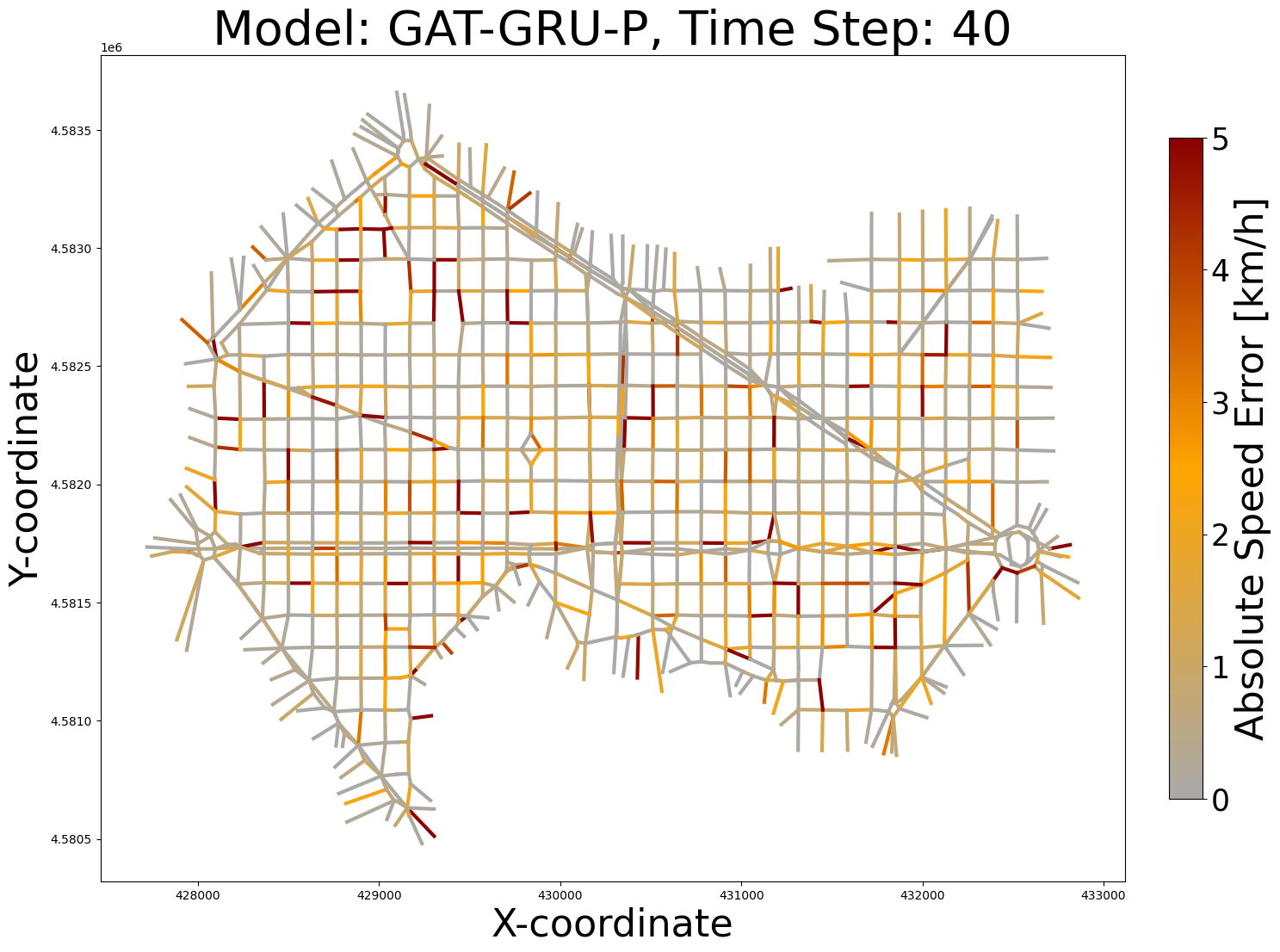}\label{fig:GAT-GRU_result}}
\caption{Absolute Error of the network at maximum network production across each model with partitioning: (a) Ground Truth (b) MFD (c) LR (d) XGBoost (e) DNN (f) GAT (g) DNN-GAT (h) GAT-GRU}
\label{fig: network speed}
\end{figure*}

\begin{figure*}[!t]
\centering
\captionsetup[subfloat]{font=scriptsize} 
\subfloat[]{\includegraphics[width=0.24\linewidth]{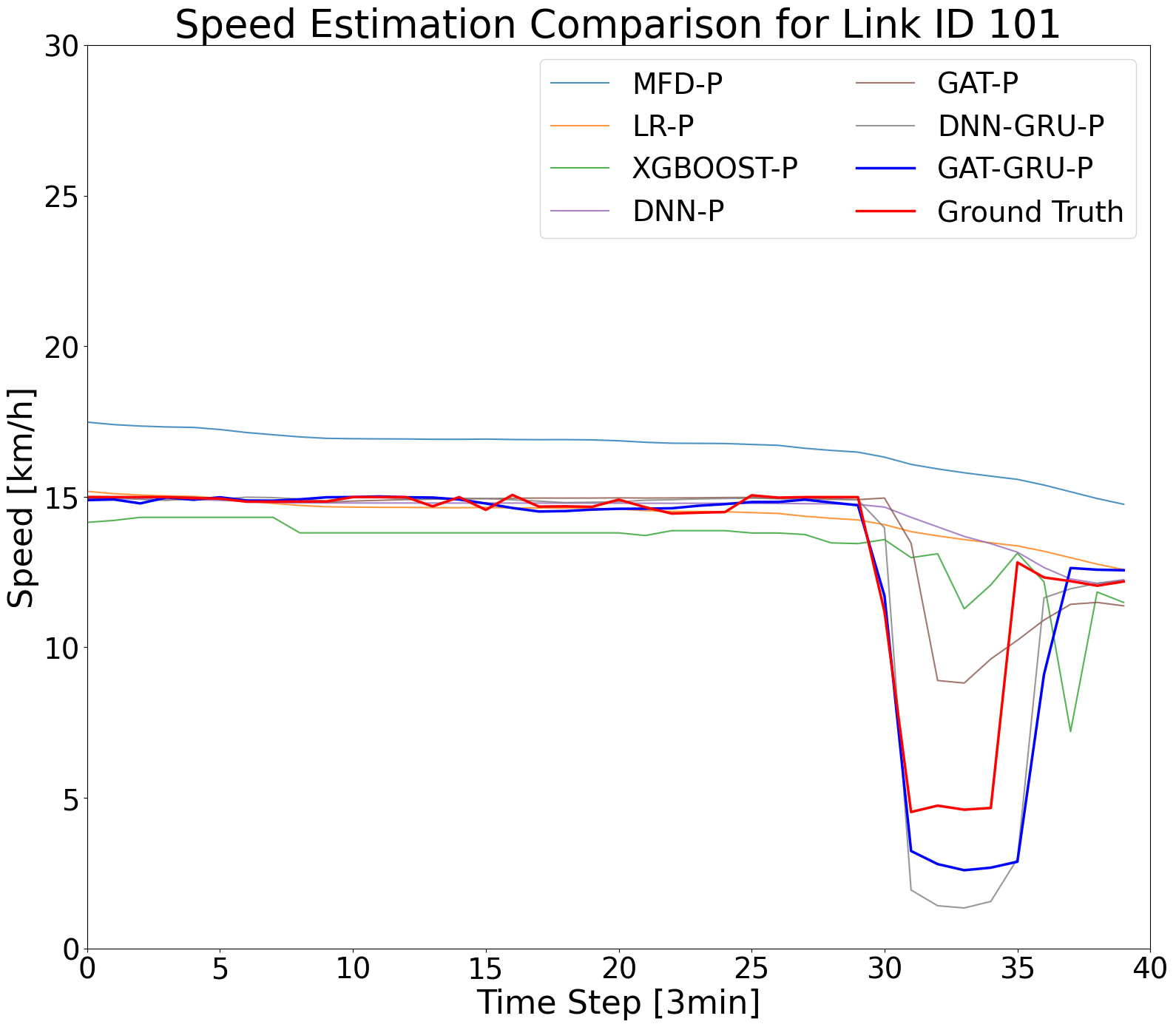}\label{fig:speed1}}
\hfil
\subfloat[]{\includegraphics[width=0.24\linewidth]{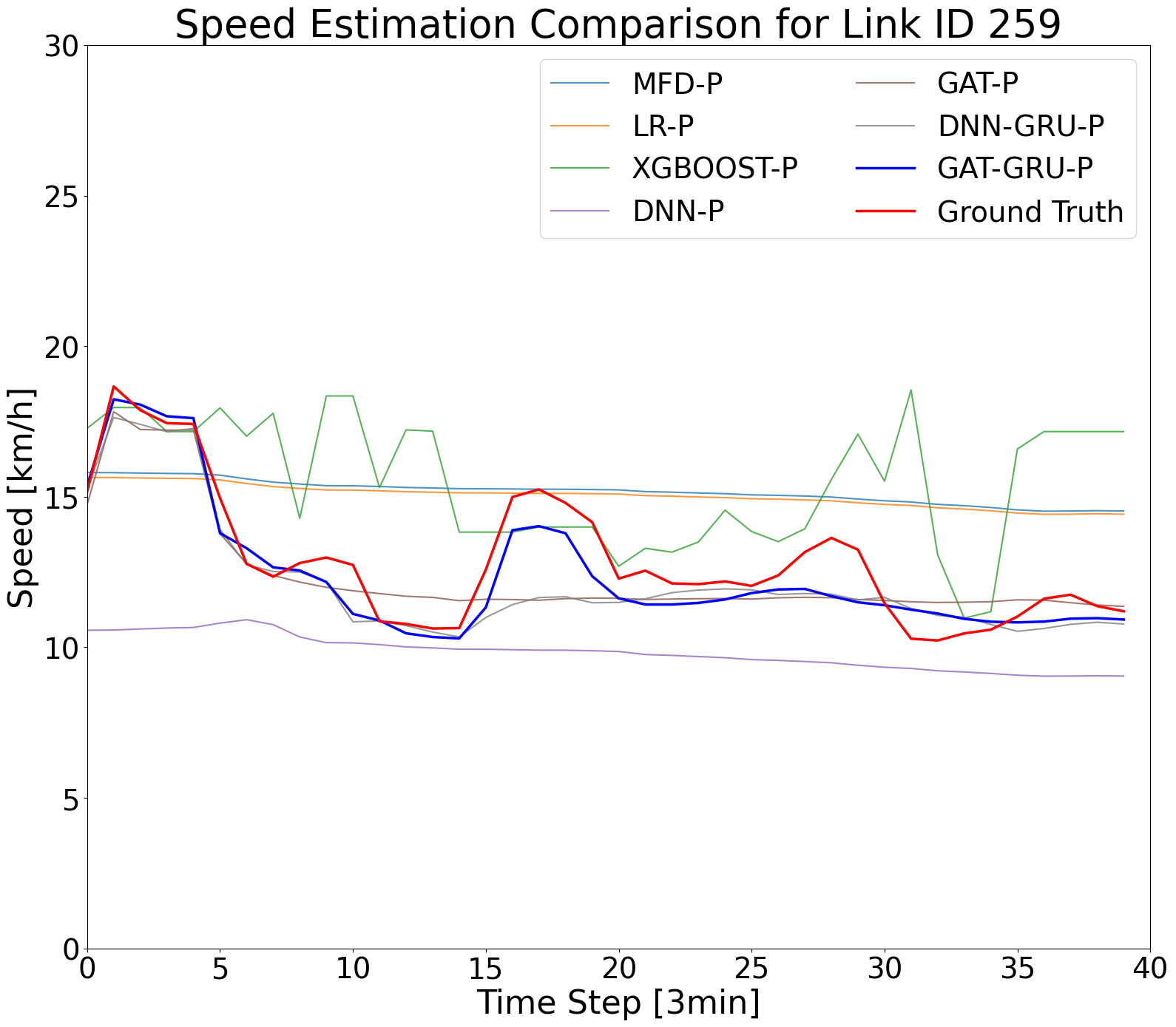}\label{fig:speed2}}
\hfil
\subfloat[]{\includegraphics[width=0.24\linewidth]{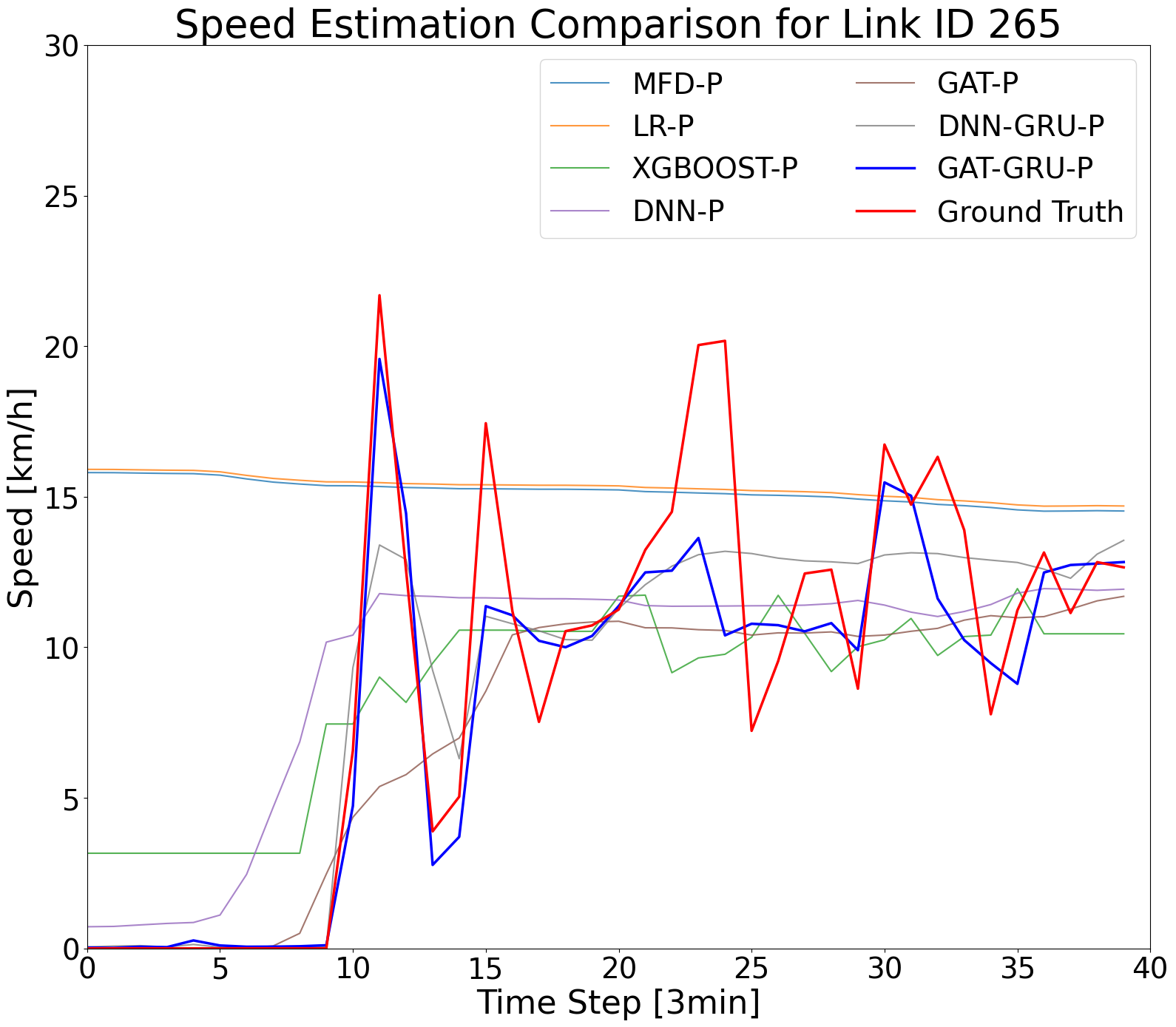}\label{fig:speed3}}
\hfil
\subfloat[]{\includegraphics[width=0.24\linewidth]{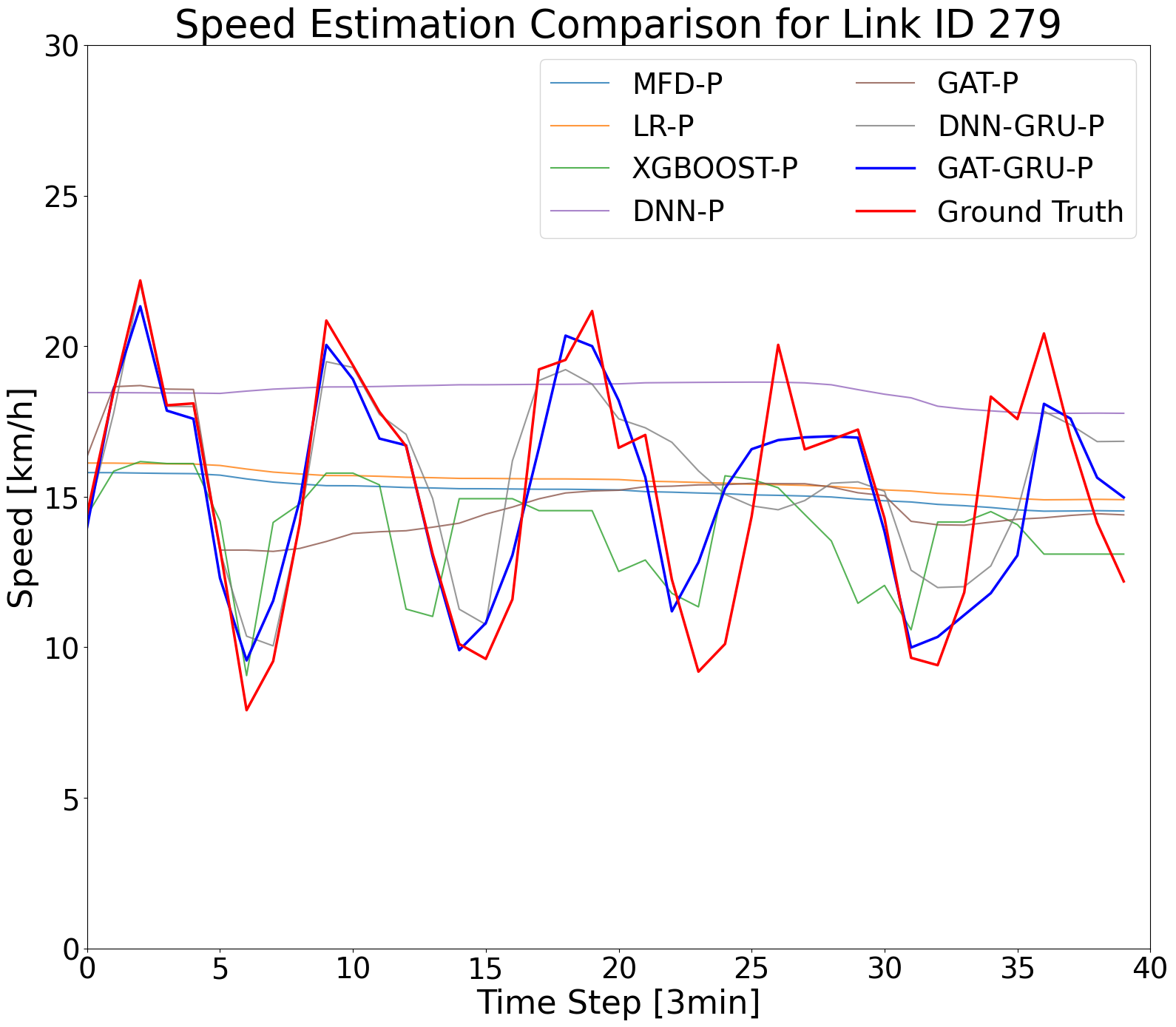}\label{fig:speed4}}
\\ 
\subfloat[]{\includegraphics[width=0.24\linewidth]{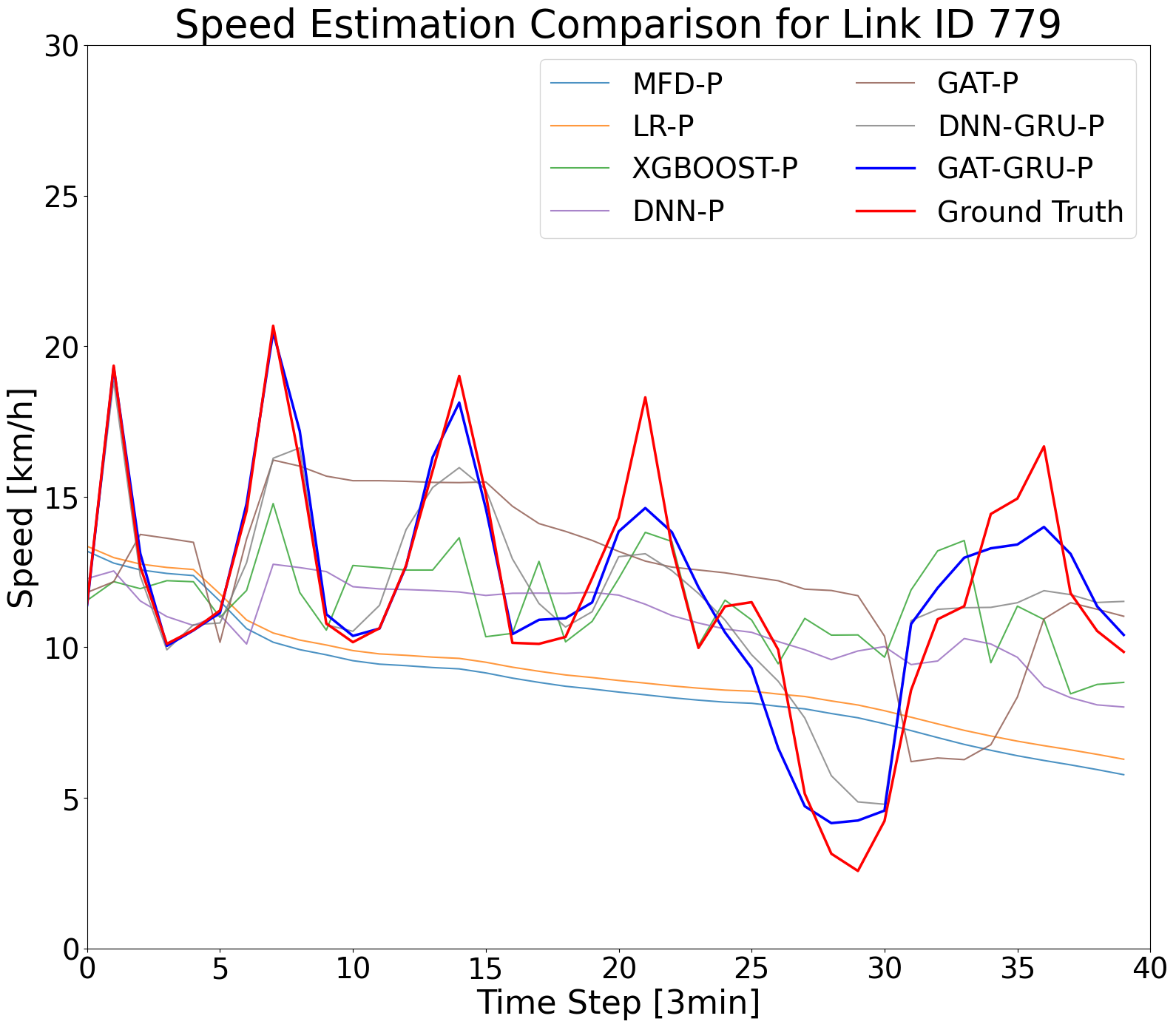}\label{fig:speed5}}
\hfil
\subfloat[]{\includegraphics[width=0.24\linewidth]{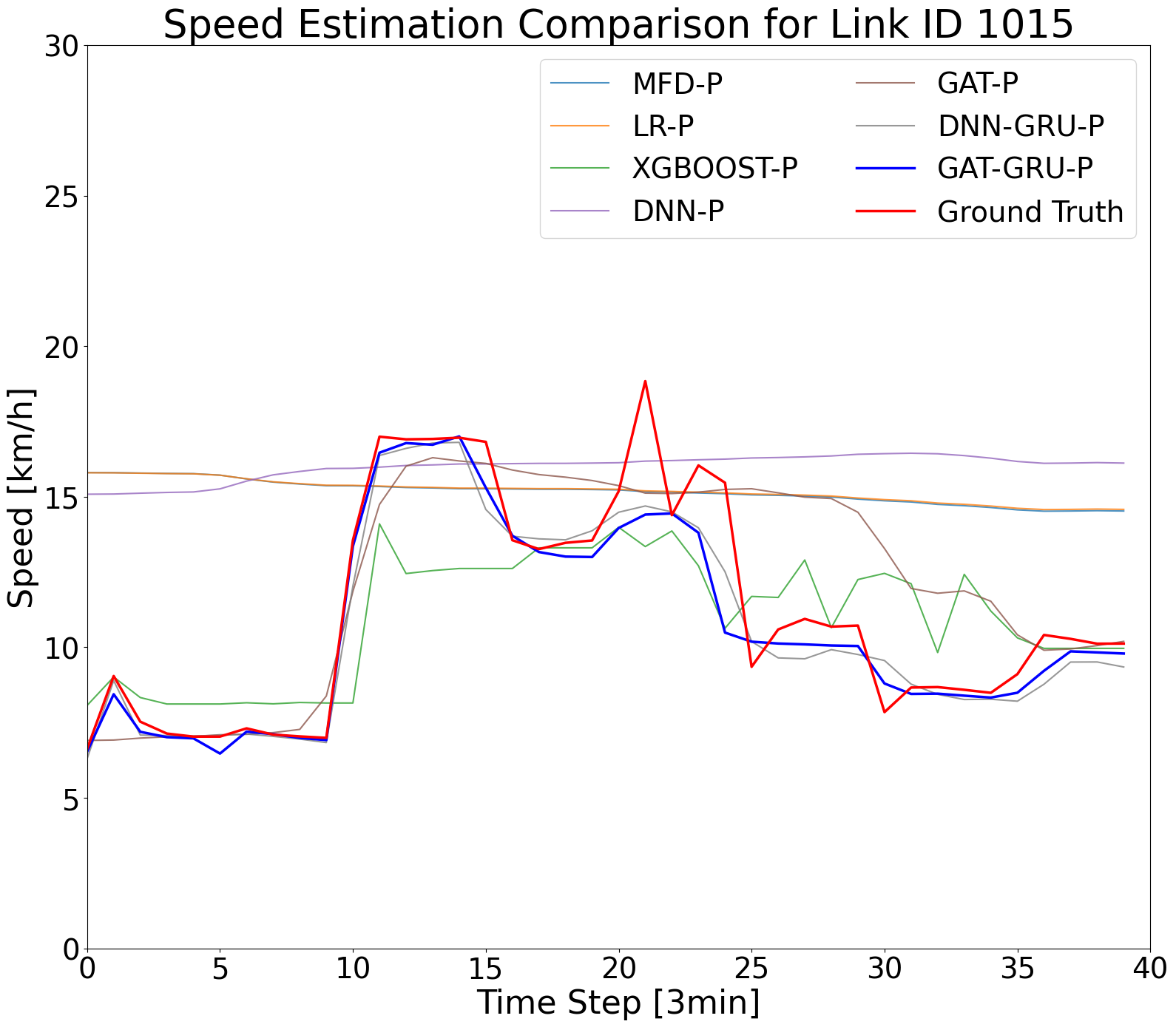}\label{fig:speed6}}
\hfil
\subfloat[]{\includegraphics[width=0.24\linewidth]{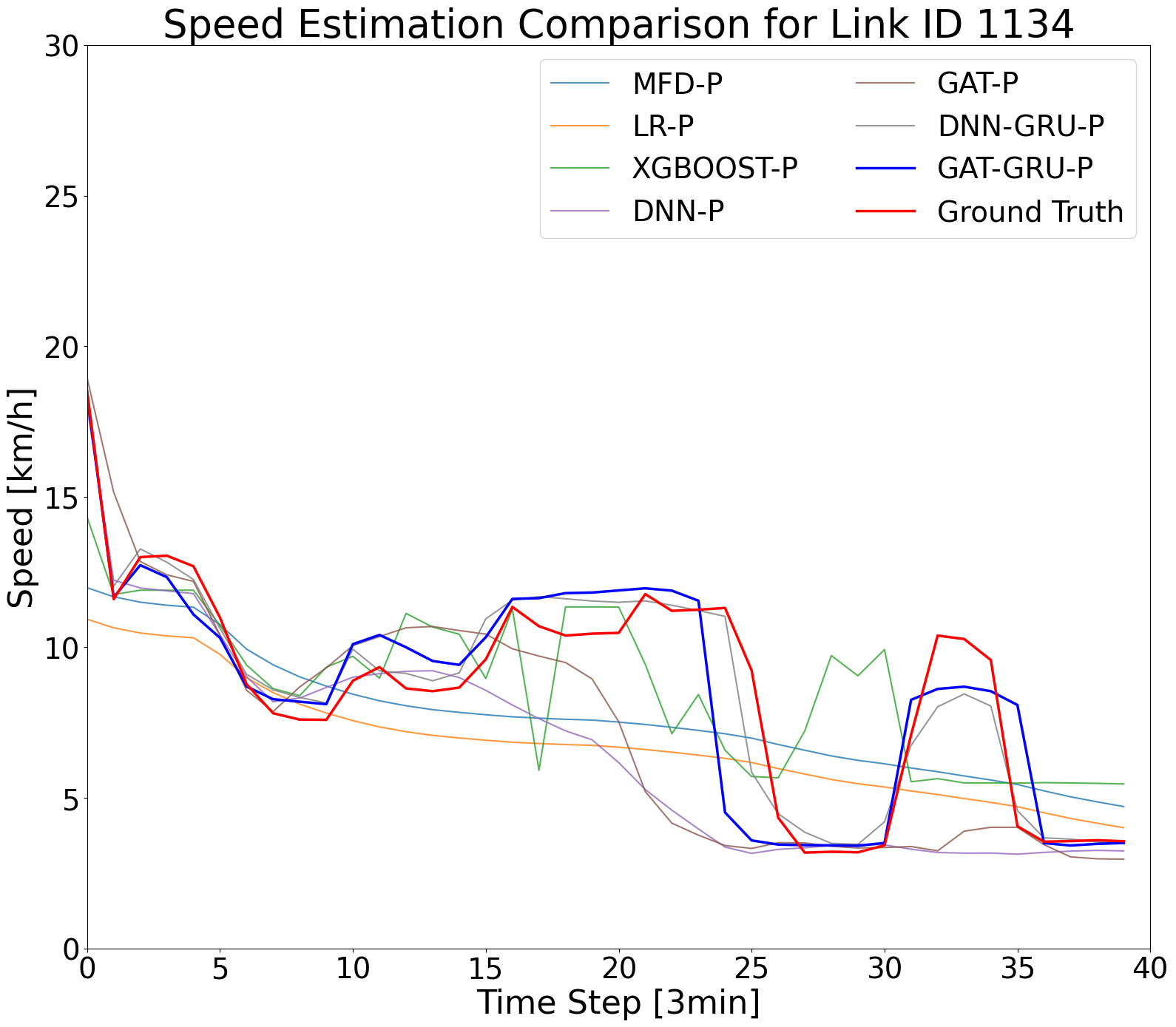}\label{fig:speed7}}
\hfil
\subfloat[]{\includegraphics[width=0.24\linewidth]{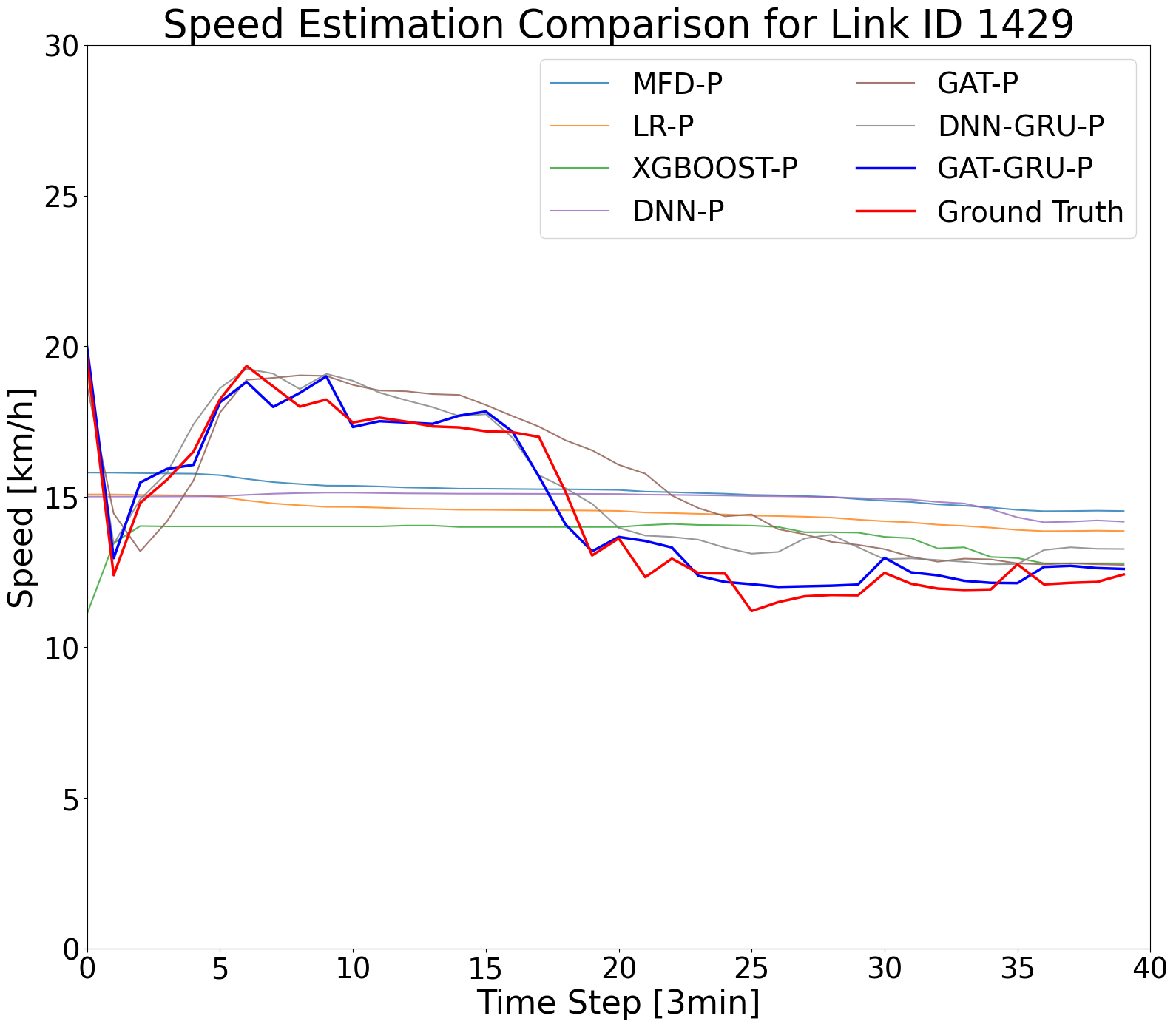}\label{fig:speed8}}

\caption{Speed profile of models at 8 different links}
\label{fig: speed profile}
\end{figure*}
To further evaluate the proposed LCF, we analyzed the error speed distributions between the MFD and the proposed GAT-GRU-P model, as depicted in Fig.~\ref{fig: Error Speed Distribution}. The MFD model, shown in gray, displays a broad error spread, indicating a high variance with substantial deviations in both directions. In contrast, the GAT-GRU-P model has a tighter distribution with a peak near zero error, demonstrating more precise and consistent speed estimations. We also visualize the results of models incorporating network partitioning from two different aspects: the absolute error of each link at peak production time and the speed profile with random road links. These visualizations are depicted in Fig.~\ref{fig: network speed} (a) - (h) and Fig.~\ref{fig: speed profile} (a) - (h). These figures clearly illustrate lower errors and closer alignment with ground truth for the models with partitioning information, particularly the proposed GAT-GRU-P model.
In Fig.~\ref{fig: network speed}, lighter colors signify lower errors. The results indicate the superior performance of deep learning models, particularly our proposed models. Similarly, in Fig.~\ref{fig: speed profile}, the proposed model shows a great matching result with the ground truth, further validating its efficacy.

\subsubsection{\textbf{LCF Robustness Evaluation}}

\begin{figure}[t]
  \includegraphics[width=0.45\textwidth]{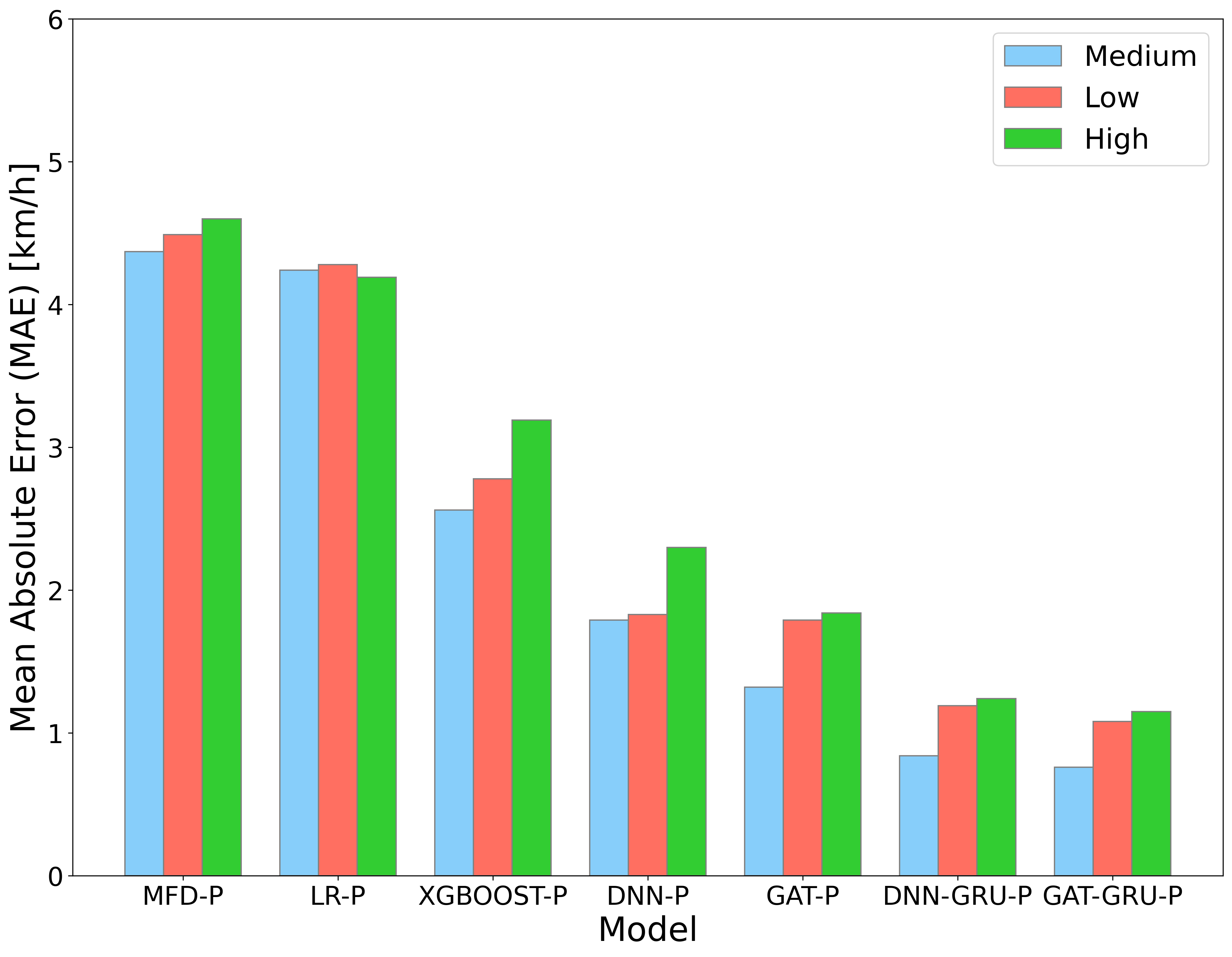}
  \centering
  \caption{Performance comparison between different demand levels across models with network partitioning}\label{fig: diff demand levels}
\end{figure}

In this section, we analyze the robustness of the proposed LCF under diverse traffic conditions, focusing on varying demand levels —low to high. This evaluation helps us understand how well the models, trained on medium-demand scenarios, adapt to changes in both lower and higher-demand volumes. The corresponding results are presented in Table~\ref{tab 2}. Consistent with the findings from the previous section, the proposed hybrid GAT-GRU model with and without network partitioning exhibits superior performance across all demand patterns, which is shown in Fig.~\ref{fig: diff demand levels}. pecifically, the MAE of the proposed GAT-GRU-P model in low and high-demand levels are 1.08 km/h and 1.15km/h, respectively. This consistency not only shows the model's effectiveness in speed estimation but also its ability to adapt to diverse traffic conditions. 

Additionally, the incorporation of the network partitioning method enhances model performance in all demand patterns. This suggests that dividing the road network into homogeneous sub-regions enhances the capture of localized traffic dynamics, thereby improving estimation accuracy. Our findings indicate that network partitioning effectively enhances speed estimation results. Regarding the impact of demand levels, the results indicate a differential impacts are observed in model performance. Specifically, under the low-demand level, the proposed GAT-GRU-P model shows an MAE of 1.08 km/h, a performance reduction of around 0.32km/h compared to the result from the medium-demand level. In contrast, under high demand, the MAE is 1.15km/h, with an approximated 0.29km/h error increase. The reason behind this is that in the case of high demand, we have more complex spatial traffic congestion patterns such as gridlocks that have not been seen in the training. While a slight performance decrease is observed in both demand patterns, the proposed model maintains a consistently high level of accuracy. The results indicate the model's robustness in various scenarios. 

\noindent

\begin{table*}[t]
\setlength{\arrayrulewidth}{0.25pt}
\centering
\caption{Performance evaluation of LCF with different range of demand levels at link level [km/h]}
\label{tab 2}
\renewcommand{\arraystretch}{1.3} 
\resizebox{\textwidth}{!}{%
\begin{tabular}{c|ccccc|ccccc|ccccc}
\hline
\noalign{\vskip-0.5pt} 
\hline
\multirow{2}{*}{Model} & \multicolumn{5}{c|}{Medium}               & \multicolumn{5}{c|}{Low}                & \multicolumn{5}{c}{High}                          \\ \cline{2-16} 
                       & MAE    & RMSE   & sMAPE(\%) & Err.Mean   & Err.STD   & MAE    & RMSE   & sMAPE(\%) & Err.Mean & Err.STD     & MAE    & RMSE   & sMAPE(\%) & Err.Mean & Err.STD \\ \hline
MFD                     & 5.67 & 6.80 & 61.78 & 0.00 & 6.80 & 5.36 & 6.49 & 48.34 & 0.00 & 6.49 & 5.65 & 6.81 & 75.94 & 0.02 & 6.81 \\
MFD-P                   & 4.37 & 5.43 & 51.61 & 0.00 & 5.43 & 4.49 & 5.47 & 42.99 & 0.00 & 5.47 & 4.60 & 5.57 & 61.14 & 0.02 & 5.57 \\
LR                      & 5.30 & 6.43 & 58.62 & 0.01 & 6.43 & 4.98 & 6.04 & 45.87 & 0.11 & 6.04 & 5.36 & 6.56 & 72.78 & -0.07 & 6.56 \\
LR-P                    & 4.24 & 5.28 & 50.77 & 0.02 & 5.28 & 4.28 & 5.25 & 41.42 & 0.10 & 5.25 & 4.19 & 5.36 & 62.80 & -0.07 & 5.36 \\
XGBOOST                 & 2.91 & 3.77 & 40.21 & 0.13 & 3.77 & 3.15 & 4.06 & 32.26 & 0.14 & 3.90 & 3.76 & 4.58 & 60.29 & -0.13 & 4.49 \\
XGBOOST-P               & 2.56 & 3.37 & 36.01 & 0.12 & 3.37 & 2.78 & 3.77 & 35.50 & 0.63 & 3.77 & 3.19 & 4.25 & 57.44 & -0.09 & 4.22 \\
DNN                     & 2.21 & 3.85 & 29.29 & 0.09 & 3.85 & 2.37 & 3.86 & 25.83 & 0.15 & 3.86 & 2.63 & 4.49 & 42.59 & 0.07 & 4.49 \\
DNN-P                   & 1.79 & 3.25 & 25.22 & 0.04 & 3.25 & 1.83 & 3.40 & 23.29 & 0.16 & 3.40 & 2.30 & 4.10 & 39.23 & 0.13 & 4.10 \\
GAT                     & 1.44 & 2.86 & 21.56 & 0.03 & 2.85 & 1.86 & 3.35 & 21.31 & 0.05 & 3.35 & 1.97 & 3.77 & 34.56 & -0.02 & 3.77 \\
GAT-P                   & 1.32 & 2.67 & 20.23 & 0.05 & 2.67 & 1.79 & 3.32 & 20.30 & 0.08 & 3.32 & 1.84 & 3.53 & 30.50 & 0.01 & 3.53 \\
DNN-GRU                 & 1.35 & 2.69 & 20.60 & 0.10 & 2.69 & 1.65 & 2.92 & 21.98 & 0.22 & 2.97 & 1.73 & 3.42 & 31.87 & -0.11 & 3.42 \\
DNN-GRU-P               & 0.84 & 1.92 & 14.45 & -0.07 & 1.92 & 1.19 & 2.27 & 14.94 & 0.24 & 2.26 & 1.24 & 2.69 & 25.87 & -0.17 & 2.68 \\
GAT-GRU                 & 1.09 & 2.42 & 17.65 & -0.02 & 2.42 & 1.44 & 2.80 & 17.50 & 0.19 & 2.80 & 1.58 & 3.37 & 30.08 & 0.10 & 3.37 \\
GAT-GRU-P               & 0.76 & 1.82 & 13.42 & 0.01 & 1.82 & 1.08 & 2.18 & 13.74 & 0.16 & 2.18 & 1.15 & 2.63 & 24.02 & -0.10 & 2.62 \\ 
\hline
\noalign{\vskip-0.5pt} 
\hline
\end{tabular}%
}
\end{table*}

\subsubsection{\textbf{LCF Applicability in Simulation}}

To evaluate the practical applicability of the proposed {\color{red}LCF} in an MFD-based simulation environment, it is crucial to accurately estimate both individual link travel times and total path travel times. While link-level estimates provide detailed insights into specific network segments, path-level travel time estimations are essential for effective traffic management. We generated 1,000 random trips with distinct origins, destinations, and start times. To determine the shortest paths based on actual travel times (using simulation-derived ground-truth speed data for each link), we employed Dijkstra's algorithm, a widely used shortest path-finding method. After establishing these paths, we applied the LCF to recalculate the travel times along the same routes. We then compared these LCF-based travel times with the ground-truth travel times to assess the accuracy and effectiveness of our proposed method.

\begin{figure}[t]
  \includegraphics[width=0.45\textwidth]{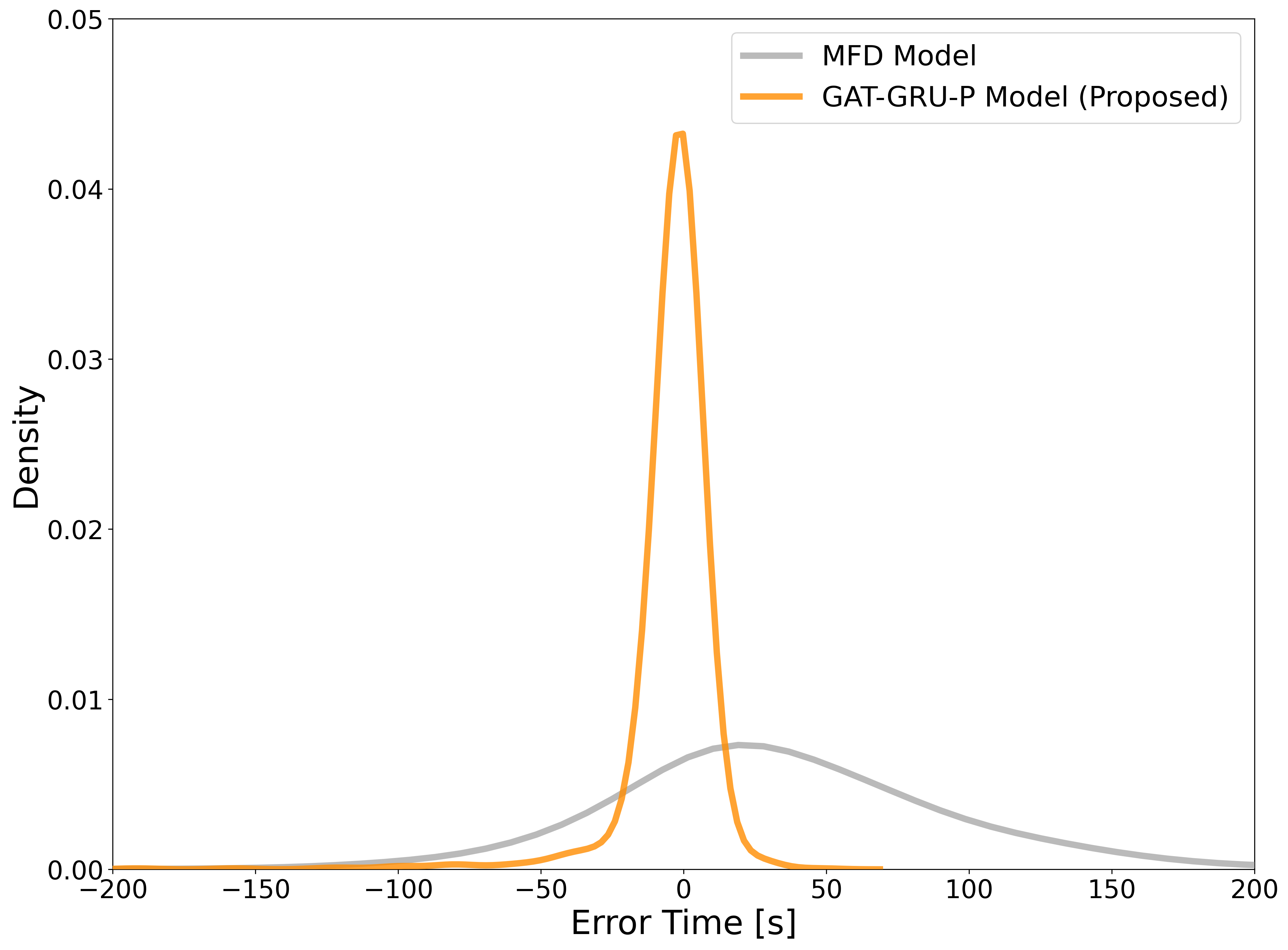}
  \centering
  \caption{Comparison of travel time error distribution between MFD and proposed model} \label{fig: Error Time Distribution}
\end{figure}

The analysis of the results is presented in Table~\ref{table 3}. 
In this experiment, the average travel time is 565s. Consistent with previous findings, models incorporating network partitioning show enhanced performance, highlighting the effectiveness of this approach in refining traffic speed estimations. Among the evaluated models, the proposed GAT-GRU-P model outperforms all other models with the lowest MAE of 9s and RMSE of 29s, indicating its applicability in the MFD-based simulators. In general, the application of the proposed LCF lowers the error of average travel time by approximately 84\%. To further analyze the effectiveness of the proposed LCF, we compare the error time distribution between MFD and GAT-GRU-P models, shown in Fig.~\ref{fig: Error Time Distribution}. The MFD model, illustrated in gray, has a wider spread on the error distribution, which implies high variance in its calculation. Its curve shows substantial tails in both the negative and positive directions, indicating frequent and varied deviations from the actual travel times. In contrast, the proposed GAT-GRU-P model exhibits a much narrower distribution that sharply peaks around the zero error time. This indicates that the model has a higher density of estimations that are close to the actual travel time, thus suggesting it is more accurate and consistent in its estimations than the MFD model. This comprehensive evaluation illustrates the potential of the proposed LCF model, especially the GAT-GRU-P model, for application in traffic management systems. The findings from this research provide a solid foundation for future research aimed at developing sophisticated solutions for traffic estimation and management, optimizing travel time, and improving the overall efficiency of urban transportation networks.



\begin{table}[t]
\caption{Results of travel time estimation at path level [s]}
\setlength{\arrayrulewidth}{0.25pt}
\centering
\renewcommand{\arraystretch}{1.15}
\label{table 3}
\resizebox{0.45\textwidth}{!}{%
\begin{tabular}{c|cccc}
\hline
\noalign{\vskip-0.5pt} 
\hline
Model     & MAE  & RMSE & Error Mean & Error STD \\ \hline
MFD       & 58   & 95   & 41         & 90        \\
MFD-P     & 53   & 90   & 30         & 86        \\
LR        & 59   & 102  & 47         & 90        \\
LR-P      & 53   & 88   & 25         & 84        \\
XGBOOST   & 52   & 84   & -31        & 80        \\
XGBOOST-P & 43   & 76   & -20        & 71        \\
DNN       & 39   & 75   & -13        & 74        \\
DNN-P     & 31   & 56   & -5         & 56        \\
GAT       & 19   & 40   & 5          & 40        \\
GAT-P     & 17   & 34   & 4          & 34        \\
DNN-GRU   & 23   & 43   & -14        & 44        \\
DNN-GRU-P & 12   & 33   & -3         & 33        \\
GAT-GRU   & 13   & 36   & -9         & 34        \\
GAT-GRU-P & 9    & 29   & -7         & 20        \\ 
\hline
\noalign{\vskip-0.5pt} 
\hline
\end{tabular}%
}
\end{table}

\noindent

\section{Conclusion} This study introduces a novel {\color{red}LCF} approach for enhancing MFD-based simulators, integrating network mean speed with its configuration to estimate the speed of each link. Leveraging {\color{red}GATs and GRUs}, our approach effectively captures the intricate spatial configurations and dynamic traffic speed of the network. Additionally, we employ a network partitioning method to further refine the accuracy of the method by dividing the network into more homogeneous regions. This advancement addresses the limitations of conventional MFD-based traffic models that typically assume uniform speed across the network while maintaining high computational efficiency.

{\color{red}In terms of contributions, this study proposes a method for link-specific speed estimation based on average speed and network configuration.} This approach can bridge the gap between loss of detail and low computation cost in MFD-based simulators. Secondly, we develop a deep learning framework combining {\color{red} GATs and GRUs}, significantly improving traffic speed estimations by accounting for both spatial configurations and temporal dynamics. Additionally, a network partitioning method is employed to divide the urban network into more homogeneous sub-regions, further refining the estimation accuracy. This novel approach ensures that the model can adapt to varying traffic conditions and network configurations, thereby increasing its applicability and effectiveness. {\color{red}Even though the LCF approach requires extensive data collection and model training, this one‐time investment enables faster, large‐scale optimization than expensive microsimulations—all within a predictable, controllable budget.}

The model's performance is evaluated on the macroscopic, queue-based SaF urban traffic simulation paradigm, which was applied in this work for the city center of Barcelona. The model performance  is assessed in terms of network partitioning visualization, LCF performance evaluation, LCF robustness evaluation, and LCF applicability in simulation. Based on the results of the network partitioning, the proposed strategy can efficiently divide the network by considering both the geographical location of each link and the average speed of the network. In the evaluation of LCF, the proposed LCF shows the best performance among other tested models, and its effectiveness in accurately estimating link speed from mean network speed is visually demonstrated. We also test the model in various demand scenarios to demonstrate its robustness in traffic volume variations. Finally, travel time calculation for random paths demonstrates the model's practical applicability and effectiveness in realistic demand-varying scenarios. 

There are several directions in which the current study could be extended to further improve the model performance. Firstly, we could explore incorporating more detailed road network information, such as more infrastructure characteristics. This information could further improve model performance. Secondly, it is necessary to test across different urban networks to validate the model's universal applicability. Furthermore, integrating the proposed LCF with MFD-based simulators for traffic management optimization represents a promising direction for further research. Such an approach may explore the potential of using the LCF in addressing complex traffic optimization problems, contributing to the development of more efficient and sustainable urban traffic systems.

\bibliographystyle{IEEEtran}
\bibliography{bibdata}

\end{document}